\documentclass{amsart}
\usepackage[dvips]{graphicx}
\graphicspath{ {.} }
\setkeys{Gin}{width=\linewidth,keepaspectratio=true}
\usepackage[latin1]{inputenc}
\usepackage{natbib}
\bibpunct{(}{)}{;}{a}{,}{;}
\usepackage[dvipsnames]{xcolor}
\usepackage{xcolor}
\colorlet{RED}{red}
\usepackage{url}
\usepackage{epsfig}
\usepackage{algorithm}
\usepackage{algpseudocode}
\usepackage{rotating}
\usepackage{lscape}
\usepackage{amsmath,amssymb}
\usepackage{numprint}
\npdecimalsign{.} 

\usepackage{mfirstuc}

\newcommand{\Xcomment}[1]{}

\begin{document}

\title[Metaheuristic for Large MWIS]
      {A Metaheuristic Algorithm for Large \\
        Maximum Weight Independent Set Problems}

\author[Y. Dong]{Yuanyuan Dong}

\address[Yuanyuan Dong]{Dallas, TX\ USA. The work was done while the author was at Amazon.com}

\email[]{njdyy03@gmail.com}

\author[A.V. Goldberg]{Andrew V. Goldberg}

\address[Andrew V. Goldberg]{Amazon.com,
East Palo Alto, CA\ USA.}

\email[]{avg@alum.mit.edu}

\author[A. Noe]{Alexander Noe}

\address[Alexander Noe]{Amazon.com,
Bellevue, WA\ USA.}

\email[]{alexander.noe@univie.ac.at}

\author[N. Parotsidis]{Nikos Parotsidis}

\address[Nikos Parotsidis]{
Google, Zurich, Switzerland. The work was done while the author was at Amazon.com}

\email[]{nickparo1@gmail.com}

\author[M.G.C. Resende]{Mauricio G.C. Resende}

\address[Mauricio G.C. Resende]{Amazon.com and Industrial \& 
Systems Engineering,
University of Washington, Seattle, WA\ USA.}

\email[]{mgcr@berkeley.edu}

\author[Q. Spaen]{Quico Spaen}

\address[Quico Spaen]{Amazon.com,
East Palo Alto, CA\ USA.}

\email[]{quico@spaen.nl}

\begin{abstract}
Motivated by a real-world vehicle routing application,
we consider the maximum-weight independent set problem:
Given a node-weighted graph, find a set of independent (mutually nonadjacent)
nodes whose node-weight sum is maximum.
Some of the graphs airsing in this application are large, having
hundreds of thousands of nodes and hundreds of millions of edges.

To solve instances of this size, we develop a new local search algorithm,
which is a metaheuristic
in the greedy randomized adaptive search (GRASP) framework.
This algorithm, which we call METAMIS,
uses a wider range of simple local search operations than previously
described in the literature.
We introduce data structures that make these operations efficient.
A new variant of path-relinking is introduced 
to escape local
optima and so is a new alternating augmenting-path local search move that
improves algorithm performance.


We compare an implementation of our algorithm with a state-of-the-art
openly available code on public benchmark sets, including some large
instances with hundreds of millions of vertices.
  Our algorithm is, in general, competitive and outperforms this 
  openly available code on
  large vehicle routing instances.
  We hope that our results will lead to even better MWIS algorithms.


\end{abstract}  
\keywords{GRASP, local search, 
maximum-weight independent set, path relinking,
heuristic, metaheuristic.
}  

\date{\today}
\Xcomment{
\thanks{Parts of this paper were written while the 
first and third authors were employed at
Amazon.com.}
}
\maketitle

\section{Introduction}
\label{s_intro}

Given an undirected
graph $G=(V,E)$, where $V$ is the set of nodes and $E$ the set 
of edges, an \textit{independent set} $S \subseteq V$ 
is a set of mutually non-adjacent nodes of graph $G$.
If each node $v \in V$ is assigned a weight $w_v$, a maximum-weight 
independent set ({\em MWIS}) of nodes $S^* \subseteq V$ is an independent set
whose sum of weights, $$W(S^*) =\sum_{v \in S^*} w_v$$ is maximum.
We denote $n = |V|$ and $m = |E|$.
\Xcomment{
Figure~\ref{2v3IS} shows a 6-node weighted graph and two independent sets.
The first independent set has three nodes and is shown in black.
It consists of two nodes of weight 4 and one of weight 5, total weight of
this independent set is 13.
The other independent set has two nodes, one of weight 7 and the other of
weight 9.
The total weight of this independent set is 16 and is maximum.

\begin{figure}[t]
	\centering
	\includegraphics[width=0.5\textwidth]{2v3IS.eps}
	\caption{Example of independent sets on a graph with six weighted nodes
		and seven edges. 
		The number on each node is the weight of the node.
		A MWIS
		of weight 16 and cardinality 2 is shown in gray while a maximum cardinality
		independent set of size 3 is shown in black.  Its weight is 13 and 
		therefore suboptimal.}
	\label{2v3IS}
\end{figure}
}

MWIS is a classical optimization problem that has been extensively
studied and has many applications \citep{But03a}.
It is one of Karp's original NP-complete problems \citep{GarJoh79a,Kar72a}.
The problem is also hard to approximate {\citep{Has99a}.

One can state MWIS as an \emph{Integer Linear Program}~-- ILP
(see Section~\ref{s:ilp}).
We can solve small MWIS problems exactly using IP solvers,
e.g., CPLEX, GUROBI, or XPRESS, or a partial enumeration algorithm,
such as the one proposed by \citet{CarPar90a}.
However, these methods do not scale to large graphs.
Over the years, heuristics have been the workhorse for solving large instances
of the maximum independent set problem 
approximately \citep{Pel09a}. 
In particular, the most successful heuristics have been the ones
based on metaheuristic algorithms, such as
GRASP \citep{FeoResSmi94a}, tabu search \citep{FirHerWer89a}, and
iterated local search \citep{AndResWer12a,nogueira2018hybrid}.

In this paper we introduce \emph{METAMIS}, a new metaheuristic algorithm
for the MWIS problem.
METAMIS is based on the
\textit{greedy randomized adaptive search procedure}~--
GRASP \citep{ResRib16a},
with \emph{truncated path-relinking}.
Our motivation is a long-haul vehicle routing (VR) application that
yields large MWIS problems, some with close to $900$ thousand nodes.
Compared to benchmark instances used in previously published work,
the VR-MWIS instances are often larger and have a very different structure.
We conduct experiments with METAMIS on MWIS instances arising in 
different applications, including on our VR-MWIS instances and on other publicly
available ones.

The paper is organized as follows.
In Section~\ref{s_hi} we give a high-level description of the algorithm.
Section~\ref{s_ds} introduces the data structure and gives low-level
implementation details.
We present experimental results in Section~\ref{s_comp} and 
make concluding remarks in Section~\ref{s_concl}.

\subsection{ILP Formulation}
\label{s:ilp}
Next we discuss several ILP formulations of MWIS.
Let $x_v$ be a binary decision variable such that $x_v = 1$ if node $v \in S \subseteq V$ and $x_v = 0$ otherwise, where $S$ is an independent set of nodes. 
A simple integer programming (IP)
formulation for selecting a maximum-weight independent 
set of nodes is
\begin{align}
\max & \sum_{v\in V} w_v x_v \notag \\
\text{subject to} & \notag \\
& x_u + x_v \leq 1, \forall \; (u,v) \in E \notag \\
& x_v \in \{0,1\}, \forall \; v \in V. \notag
\end{align}
The objective is to maximize the sum of the weights of the nodes selected to be
in the independent set $S$.
The constraints guarantee that all selected nodes are mutually
non-adjacent, i.e. for all edges in the graph, at most one node can be
in the independent set.
Stronger formulations add \textit{clique inequalities} 
\citep{Pad73a} to the above formulation.
For a 2-clique, or clique of size 2, denoted by $C_2$ we have 
$x_u + x_v \leq 1$, for all $(u,v) \in E$.
A 3-clique ($C_3$) inequality is $x_u + x_v + x_t \leq 1$, for all
triangles, i.e. $u,v,t \in V$, such that
$(u,v) \in E$, $(u,t) \in E$, and $(v,t) \in E$. 
In general, for any clique $Q$, we have a constraint 
\begin{displaymath}
\sum_{ v \in Q} x_v \leq 1.
\end{displaymath}
Let $C_2, C_3, \ldots, C_k$ be, respectively, the sets of 2-clique, 
3-clique, $\ldots$, and  $k$-clique inequalities.
The clique IP formulation for maximum weight independent set is:
\begin{align}
\max & \sum_{v\in V} w_v x_v \notag \\
\text{subject to} & \notag \\
& C_2, C_3, \ldots, C_k, \label{ilp} \\
& x_v \in \{0,1\}, \forall \; v \in V. \notag
\end{align}
Other tight formulations for independent set can be found in \citet{BucBut14a}.

In the \emph{linear programming (LP) relaxation} 
of the problem, we allow fractional solutions:
$x_v \in [0,1]$.
The relaxed problem is much simpler and can be solved by an LP solver
in reasonable time.
We use an optimal solution to the LP relaxation to improve the performance
of our local search algorithm.

The fact that for some problems an LP relaxation can help to find a good
feasible solution is well-known.
For example, for a class of IP problems, randomized rounding of the optimal
relaxed solution yields a solution with a provably good approximation
ratio \citep{RT87}.
However, for MWIS, a rounded solution may not be feasible.
Our use of the relaxed solutions can be viewed as another way of extracting
information about good integral solutions from the relaxed solution.

Note that as long as all 2-clique inequalities are present, for any set of
additional clique inequalities, the set of feasible integer solutions to the
corresponding IP is the same as for the simple formulation.
Additional clique constraints make the linear programming (LP) relaxation
stronger.

Although our algorithm is a general-purpose heuristic, our motivation comes
from vehicle routing (VR) 
\citep{DonGolNoeParResSpa21a,DonGolNoeParResSpa21b}.
A variant of our
algorithm takes advantage of the application-specific structure for performance and solution quality.
In this application, we have a good initial solution.
One can use this solution to warm-start our MWIS algorithm.
Our experiments show how much the warm-start improves solution quality.


\section{High Level Description}
\label{s_hi}

The MWIS algorithm is an iterative local search algorithm based on
the {\em Greedy Randomized Adaptive Search Procedure (GRASP)}
metaheuristic, which is
a general metaheuristic for combinatorial optimization
\citep{FeoRes89a,FeoRes95a,ResRib16a}.
The algorithm also uses {\em path relinking} to escape local optima
\citep{LagMar99a,ResRib16a}.

\begin{algorithm}
\begin{algorithmic}[1] \Procedure{MWIS}{$G = (V, E, w),$ maxTime, $S_0$}
    \State $S \leftarrow \text{localSearch}(G, S_0)$
    \State $\mathcal{ES} \leftarrow \{\}$ \Comment{Empty set of elite solutions}
    \State $\mathcal{ES}.\text{add}(S)$
    \While{$t \leq \text{maxTime}$}
    \State $S_G \leftarrow \text{findRandomizedGreedySolution(G)}$
    \If{$\text{LsBeforeRelinking}$} \Comment{Optional local search}
    \State $S_G \leftarrow \text{localSearch}(G, S_G)$
    \EndIf
    \State $S_e \leftarrow \mathcal{ES}$.randomEliteSolution$()$
    \State $S' \leftarrow \text{pathRelinking}(G, S_G, S_e)$
    \State $S' \leftarrow \text{localSearch}(G, S')$
    \State $\mathcal{ES}.\text{tryToAddAndEvict}(S')$ \Comment{Add solution to elite set, if full evict similar solution of lesser value (or don't insert if no worse elite solution exists)}
    \EndWhile
    \State \textbf{return} $\mathcal{ES}$.bestSolution()
    \EndProcedure
\end{algorithmic}
\caption{Algorithm Overview \label{alg:overview}}
\end{algorithm}

Figure~\ref{alg:overview} gives a high-level view of the algorithm.
In addition to the graph, the input to the algorithm includes a stopping
criterion, e.g., a time limit,
and an initial solution.
In our application we have a good initial solution.
When no such solution is available, one can find a solution using
the randomized greedy algorithm described later in this section.
The algorithm applies local search to improve the initial solution and
enters the main loop.
At termination of the local search procedure, we are at a local optimum.

The algorithm maintains a set of \emph{elite} solutions $\mathcal{ES}$,
which are the best solutions we have seen so far.
We add a solution to $\mathcal{ES}$ immediately after a local search,
so the elite solutions are always locally optimal.
At each iteration of the loop, we first attempt to escape the local optimum
corresponding to the elite solution.
In the process, we can decrease the objective function.
To escape a local optimum, we first find a randomized greedy solution $S_G$.
Optionally, we apply local search to improve $S_G$.
Then we apply path relinking to $S_G$ and a random elite solution from $\mathcal{ES}$ to find
a new solution $S'$.
Then we apply local search to improve $S'$, and update $S^*$ if
we find a better solution.

In our experiments, we omit the optional call to local search immediately after
path relinking because this variant of the algorithm seems to work
better for the VR-MWIS instances.
We also set the size of the elite set $\mathcal{ES}$ to $1$,
so we only retain the best solution.
This setting works best for the VR-MWIS instances.
For other problem families, different parameter choices were found
to work better
{\citep{Kummer2020,KumResSou2020a}}.

\subsection{Greedy Algorithm}
\label{s_gr}
We construct a {\em greedy solution} $S$ as follows.
To initialize $S$, we add to it all zero-degree nodes
and delete these nodes from the graph.
Next, we define $\eta(v) = w(v) / \mbox{\rm degree}(v)$
and create a list of the nodes sorted by $\eta(v)$.
Initially the list contains all nodes.
At every step, we remove node $v$ with the largest $\eta(v)$ from the list,
add $v$ to $S$, and remove neighbors of $v$ from the list.
We terminate when the list is empty.

To {better explore the solution space},
we need diverse solutions, so we {\em randomize}
greedy algorithm as follows.
Instead of removing the node with the largest $\eta(v)$ value,
we remove a random node from the set of $k$ nodes with the largest values.
We set $k$ to {be} a fraction of $n$, e.g., 10\%.

The intuition behind the greedy algorithm is that a node with a large $\eta(v)$
value is more likely to be in an independent set of high weight.
However, as we add nodes to $S$ and delete these nodes and their
neighbors from the graph, residual node degrees change.
This motivates the {\em {adaptive}} greedy algorithm.

The adaptive algorithm maintains node degrees in the graph with nodes in $S$
and their neighbors deleted.
We maintain a priority queue of nodes in the graph.
If we delete one or more neighbor of a node $v$, do the following.
If degree of $v$ becomes zero, we add $v$ to $S$ and delete it from the graph.
Otherwise, $\eta(v)$ increases, and we use increase-key operation to update
the priority queue.
Assuming that the increase-key operation takes constant time and other
priority queue operations take logarithmic time (e.g., we use Fibonacci
heaps~\citep{FT86}),
the complexity of the greedy algorithm is $O(m + n\log n)$.

In practice, for large graphs the {adaptive} algorithm is expensive.
To mitigate the problem, we use it in the first local search (line 2
of Algorithm~\ref{alg:overview}), which results in a substantial improvement
over using the deterministic algorithm.
However we use the latter algorithm in the main loop and rely on path
relinking to improve solution quality.
We do not randomize the {adaptive} greedy algorithm as we only use it once,
in the first local search.

\subsection{Local Search}

\begin{algorithm}
    \begin{algorithmic}[1]
        \Procedure{LocalSearch}{$G = (V, E, w), S, \text{numIterations}$}
        \State $i \leftarrow 1$
        \State $S^* \leftarrow S$
        \While{$i \leq \text{numIterations}$}
        \State $S_i \leftarrow \{\}$ \Comment{Empty solution}
        \While{$w(S_i) < w(S)$} \Comment{Repeat until no improvement is found}
        \State $S_i \leftarrow S$
        \State $S \leftarrow$ starOneMoves$(G, S)$
        \State $S \leftarrow $ AAPMoves$(G, S)$
        \State $S \leftarrow $ oneStarMoves$(G, S)$
        \If{$w(S_i) < w(S)$}  {\bf return} \Comment{Solution improved}
        \EndIf
        \State $S \leftarrow $ twoStarMoves$(G, S)$
        \EndWhile
        \If{$w(S) > w(S^*)$}
        \State $S^* \leftarrow S$
        \State $i \leftarrow 1$
        \Else
        \State $S \leftarrow $ perturb$(S)$
        \EndIf
        \EndWhile
        \EndProcedure
    \end{algorithmic}\caption{Local Search Procedure\label{alg:localsearch}}
\end{algorithm}

The local search procedure, outlined in Figure~\ref{alg:localsearch}, repeatedly
performs local moves with positive gain.
We aim to find positive gain (\emph{improving}) moves 
until we reach a local optimum,
and then we perform a random perturbation of the solution.
If we find an improving move, we apply it immediately.
We use a subset of $(x,y)$ moves and
{\em alternating augmenting path moves (AAP-moves)}.
While the $(x,y)$ moves have been studied previously, the AAP moves are new.
We describe the moves at {a} 
high level in this section, {and} give a detailed
description in Section~\ref{s_ds}.

An $(x,y)$ move removes 
$x$
nodes from the solution and adds 
$y$
nodes
to it
while maintaining solution independence.
We use $*$ instead of $x$ or $y$ to denote an arbitrary positive integer.
Note that the number of applicable moves increases significantly as $x$ and $y$
increase.
Previous algorithms used $(x,y)$ moves for small values of {$x$ and $y$}.
In particular, the algorithm of {\citet{nogueira2018hybrid}} uses $(*,1)$
and $(1,*)$ moves.
Our algorithm uses $(*, 1)$, $(1, *)$, and 
$(2, *)$
 moves.
The number of 
$(2, *)$
moves is large.
We use data structures and operation ordering that {make} improving moves
more likely, {which makes our algorithm more efficient}.

If an $(x,y)$ move renders $S$ non-maximal, we add nodes without a neighbor
in $S$ to the independent set in random order until $S$ becomes maximal.
Note that through this update sequence, $S$ remains an independent set.

A $(*,1)$ move is the simplest.
It inserts a single node $u$ into the current solution $S$ and removes its
neighbors from $S$.
Procedure starOneMoves$(G, S)$ applies the $(*,1)$ moves while no such improving move.

A $(1, *)$ move removes {a node $v$} from $S$ and adds to $S$
an independent subset $I$ of the nodes whose only neighbor in $S$
before the removal is $v$.
Usually one has multiple choices of independent sets to add.
A good heuristic is to add a maximum weight set of the neighbors that
maintains independence.
This is done when the number of neighbors is small (at most seven in our
experiments).
We use a naive recursive algorithm:
Pick a node $u$
 in the neighborhood and recursively solve 
two subproblems.
The first subproblem results by 
adding $u$ to $S$
and
deleting its neighbors from the graph.
We get the second subproblem by 
deleting $u$ without adding it to $S$.
The better of the two corresponding solutions is returned.
When the neighborhood is big, a variant of the greedy algorithm of
Section~\ref{s_gr} is used.
In this case, however, we found that it is {better} to process nodes
{in descending order} of their weight 
$w(v)$
{and not of $\eta(v) = w(v)/\mbox{\rm degree}(v)$}.
Procedure oneStarMoves$(G, S)$ applies the $(1,*)$ moves while no such improving move.

A $(2, *)$ move removes two nodes, $u$ and $v$, from $S$
and adds to $S$ an independent subset $I$ {of
the nodes} whose only neighbors in $S$ before the removal is $u$, or $v$,
or both $u$ and $v$.
Generally, this set is significantly larger than the
corresponding set for the $(1, *)$ moves, and the recursive operation
used for the $(1, *)$ moves is too expensive.
One could use greedy addition, but in our experiments a random addition,
that adds to $S$ a random node from $I$ that has no neighbors in $S$,
{was better}.
We also tried biased random addition that picks a node with probability
proportional to its weight, but uniformly random selection was better.
Procedure twoStarMoves$(G, S)$ applies the $(2,*)$ moves until it finds an improving
move or there is no improving $(2,*)$ move. Note that unlike the corresponding
procedures for other moves, twoStarMoves exits as soon as it finds an improving move.

Our idea for AAP moves comes from matching algorithms~\citep{Edmonds65},
although we use a somewhat different definition.
Given an independent set $S$, we define {an AAP} $P$ as follows.
Let $I = S \cap P$ and $O = P - S$ be nodes of $P$ that are in and out
of $S$, respectively.
\begin{enumerate}
\item
  if $v \in I$, then the neighbors of $v$ on $P$ are in $O$,
\item if $v \in O$, then the neighbors of $v$ on $P$ are in $I$,
\item
  if we {\em flip} the path, i.e., set $S = S - I + O$, $S$ remains an
  independent set.
\end{enumerate}
An AAP move finds an alternating augmenting path, flips it, and looks
at the change in $w(S)$.
If the change is positive, we accept the AAP move; otherwise we reject the move.
For efficiency, we apply a limited number of AAP moves.
We postpone the details of finding the AAP moves until Section~\ref{s_ds}.
Procedure AAPMoves$(G, S)$ applies the AAP moves while no such improving move.

During an execution of the algorithm, most local search moves do not
improve solution quality and thus do not change the solution.
Note that complexity of evaluating $(2,*)$ moves is significantly higher than those for
the other moves.
Our local search repeatedly applies starOneMoves, AAPMoves, and oneStarMoves
procedures while these procedures find improving moves.
If we find an improving move, an immediate application of these procedures
may find additional improvements due to neighborhood changes, so we iterate.
Only when these procedures fail to find improving moves we call twoStarMoves.
If twoStarMoves fails to improve the solution, we perform a random perturbation.

The perturbation adds a small set of random nodes to $S$ and removes
their neighbors.
After perturbing, we resume local search.
The local search algorithm terminates if there has been no improvement to the
best solution after a predefined number of iterations.

\subsection{Using the Relaxed LP Solution} \label{s_lp}
In our VR application, we use clique information and get a relaxed LP solution
to the relaxed problem~\eqref{ilp}.
The solution assigns a value $x_v \in [0,1]$ to each node $v$.
We use these values to bias random node selection in the perturbation step
of the local search.
When performing a random perturbation in Algorithm~\ref{alg:localsearch},
we add a node $v$ to the solution with probability proportional
to $x_v + \epsilon$.
Here $\epsilon$ is a positive value (set to $\epsilon = 0.005$)
that ensures that each node can be picked, even if $x_v = 0$.
This guides the local search by biasing route selection toward nodes
with higher fractional relaxed solution value.
The idea is that a node with a high value in the relaxed solution is
more likely to be part of a good solution for the MWIS.
Using prefix sums we can pick a random node in time $\mathcal{O}(\log{|V|})$:
We draw a random floating-point number $z \in [0,\sum_{v \in V} x_v)$ and
use binary search on the prefix sum array to pick a node such that the
sum up to but excluding the node is less than $z$, and the sum up to and
including the node is greater or equal to $z$.

\subsection{Adaptive Path Relinking}

Path relinking is a technique for escaping local optima.
We discuss this technique in the context of MWIS, assuming that the local
search moves are symmetric: the reverse of a move is a valid move.
Define an undirected graph associated with the search space MWIS, 
where the nodes correspond to feasible solutions and the edges 
correspond to local search moves that transform the solution corresponding
to the tail of the edge to the solution corresponding to the head.
A path in this graph corresponds to a sequence of the moves that transform
the solution at one end of the path into a solution at the other end.
Note that the moves need not improve the objective function value.
The underlying assumption of path relinking is that if the 
{end-points} of a path
correspond to high quality solutions, then the path will contain previously
undiscovered high-quality solutions.

For our local search, given two solutions $S$ and $T$, we can transform $S$
into $T$ as follows.
Initialize $S' = S$.
At every step, we do either a 
$(*,1)$ move or a $(1,*)$ move.
In the former case, pick a node $v \in T - S'$, add $v$ to $S'$, and
remove neighbors of $v$ from $S'$.
In the latter case, pick a node $v \in S',\; v \not \in T$ and remove
$v$ from $S'$.
Let $N(v)$ denote the set of neighbors of $v$.
Then we iterate over nodes $u$ in $N(v) \cap T$.
If $N(u) \cap S' = \emptyset$, we add $u$ to $S'$.

For large graphs, finding good solutions is expensive.
Instead of combining two good solutions, we apply path relinking to combine
the randomized greedy solution $S_G$ with the current best solution $S^*$,
which is locally optimal.
While $S^*$ is a good solution, $S_G$ 
{may not be} good, and the solutions
on the path far from $S^*$ are {usually} not good {either}.
We modify path relinking so that it examines only a prefix of the path
close to $S^*$.
The prefix is small enough so that the solution quality remains good,
yet big enough so that the subsequent local search will not end up with
a locally optimal solution equivalent to $S^*$.
{This an {adaptive} variant of the {\em truncated greedy path relinking}
described in \citet{ResMarGalDua10a}}.

The first modification is to choose the node $x$ to add to $S$ or to remove
from $S$ greedily.
We pick a node that maximizes the weight of the solution we get after the move.
A second modification is to do a truncated
path relinking:
we stop the process after a certain number of steps, which we adjust adaptively.
We start with a small limit on the number of steps and increase the limit
if the algorithm gets stuck in a local optimum of weight $w(S^*)$.

More precisely, we stop if the weight factor $f = w(S)/w(S^*)$ is smaller
than a pre-defined limit, or we performed more than $c_n$ negative gain
changes, or made more than $c_p < c_n$ positive gain moves.
Positive gain moves help us escape local optima, as they will not be undone
by the $(1, *)$ moves of the subsequent local search.
We set the initial parameter values as follows:
$f = f_0 = 0.9998$, $c_n = c_{n0} = 1.0$ and $c_p = c_{p0} = 0.1$.
If $w(S) = w(S^*)$ after local search on $S$, we multiply
$f' = f \cdot 0.9998$, $c_n' = c_n \cdot 1.5$ and $c_p = c_p \cdot 1.5$.
If $w(S) \neq w(S^*)$, we reset the limits to their initial values.
If the local search algorithm finds a solution $S$ with $w(S) = w(S^*)$,
we check whether the solutions are equivalent and terminate local search if
they are.
Two solutions are equivalent if they are the same or if we can transform one
into another by zero-gain $(1, *)$ and $(*, 1)$.


\section{Data Structures and Optimizations}\label{s_ds}

As we are interested in solving problems on graphs with hundreds of millions
of edges, the choice of data structures is important for the efficiency
of the algorithm.
When making trade-offs between performance on sparse and dense graphs
we favor the former because our motivating application yields relatively
sparse graphs.

Several of our data structures use sets of objects.
We use a representation of sets based on hashing.
This representation allows constant time addition, deletion, and
membership query, and linear time iteration over all set elements.
We also assume that if we add an element to the set that already contains
the element, the set does not change.
Similarly, if we delete an element not in the set, the set does not change.

\subsection{Input Graph}

The input graph is static: it does not change throughout the execution.
We assign to the nodes of the graph integer IDs from $[0,\ldots, n-1]$
and place them in an array, with node $i$ in position $i$.
Each node has an array of edges {incident} to it.
This places the edges {incident} to a node in contiguous memory locations,
assuring that a common operation of scanning an edge list has a good memory locality.
We sort edges by IDs of the head node.
This allows us to do neighborhood queries (e.g., ``Is $v$ in $N(u)$?'')
in time logarithmic in the degree of $u$ using binary search.

Note that using sets to represent neighborhoods would give constant
neighborhood queries and linear time edge list scan.
However, the constant factors, both in terms of running time and
memory consumption, associated with hashing are large.
In addition, we lose the locality in edge list scans.
For graphs arising from our motivating application, the array-based
implementation is significantly faster than the one based on sets.

\subsection{Interstate Graph}\label{s_is}
The \emph{interstate graph} makes the local search operations more efficient.
To describe this graph, we need a few definitions.

For a node $u \in S, (u,v) \in E$, we say that $v$ is a {\em 1-tight neighbor}
of $u$ if $N(v) \cap S = \{u\}$ {\citep{AndResWer12a}}.
Note that if we remove $u$ from $S$, we can add to $S$ any 1-tight
neighbor of $u$.

Two nodes $u,v \in S$ are \emph{mates} if for at least one node $w \not\in S$,
$w$ has exactly two neighbors in $S$: $N(w) \cap S = \{u,v\}$.
We call the node $w$ a \emph{2-tight neighbor} of $u$ and $v$.
We say that $w$ is a 2-tight neighbor of $u$ if $u$ has a mate $v$
such that $w$ is a 2-tight neighbor of $u$ and $v$.
If we delete $u$ and $v$ from $S$, we can replace them by an independent set of
the union of three sets: the set of the 1-tight neighbors of $u$,
the set of 1-tight neighbors of $v$, and the set of the shared
2-tight neighbors of $u$ and $v$.

Our main data structure is the \textit{interstate graph}
$G_{\mathit{IS}} = (V, E_{\mathit{IS}}, w)$.
For $G_{\mathit{IS}}$, the nodes and node weights are the same as in the input
graph $G$.
The edge set $E_{\mathit{IS}}$ is changed dynamically depending on the nodes in the
current independent set $S$.
$E_{\mathit{IS}}$ has three types of edges:
\begin{enumerate}
\item $e = (u,v) \in E$, where $u \in S$ and $v$ is a 1-tight neighbor of $u$;
\item $e = (u,w) \in E$, where $u \in S$ and $w$ is a 2-tight neighbor of $u$;
\item $e = (u,v)$, where $u, v \in S$ are mates.
\end{enumerate}
We represent the three edge types separately.
\begin{enumerate}
\item
  For every $u \in S$, we represent its 1-tight neighbors as sets.
  For $v \not \in S$ that is a 1-tight neighbor of $u$ we add the
  \emph{1-tight} edge $(v,u)$.
\item
  For every pair of mates $u$ and $v$, we maintain a set of 2-tight neighbors
  of $u$ and $v$.
  For every 2-tight neighbor $w \not \in S$, we add the pair of \emph{2-tight}
  edges $(w,u)$ and $(w,v)$.
\item
  For every node $v$ in $S$, we maintain a set $M_v$ of its mates.
  Every mate $w \in M_v$ corresponds to a mate edge $(v,w)$.
\end{enumerate}

\subsection{Efficient Implementation of $(x, y)$ Moves}\label{s_xy}
In this section we show how to efficiently {implement $(x,y)$
moves} using the interstate graph and two additional optimizations,
one for the $(1,*)$ moves and another for $(*,y)$ moves.
We discuss maintenance of the interstate graph in Section~\ref{s_is}.

To implement 
$(*,1)$ operations efficiently, we use an idea
{from \citet{nogueira2018hybrid}}.
For every $u \not \in S$, we maintain a value
$$\Delta(u) = w(u) - \sum_{v \in S\cap N(u)} w(v)$$ to speed
up the 
$(*, 1)$ moves.
Such a move is an improving move when $\Delta(u) > 0$.
We keep a set $S^+$ of the nodes $u$ with $\Delta(u) > 0$.
Note that for an efficient implementation of 
$(*, 1)$ moves,
we need to update the vector $\Delta(\cdot)$ and the set $S^+$.
We do this as follows.
Every time we add a node $u$ to $S$, we remove $u$ from $S^+$.
Then for each $v \in N(u), \; v \not \in S$, we set
$\Delta(v) = \Delta(v) - w(u)$.
Every time we remove $u$ from $S$, we scan the edge list of $u$ and compute
$\Delta(u)$.
If $\Delta(u) > 0$, we add $u$ to $S^+$.
Also during the scan, for every neighbor $v$ of $u$ such that $v \not \in S$,
we increase $\Delta(v)$ by $w(u)$, and if $\Delta(v)$ becomes positive, we
add $v$ to $S^+$.
We have an improving 
$(*, 1)$
 move if and only if $S^+$ is non-empty.
In this case, we can pick a node $u$ from $S^+$ and apply the $(*, 1)$
move to it.

Since for every $u \in S$ we maintain a set of its 1-tight neighbors
as a hash set, we can efficiently run the recursive or the greedy algorithm
described in Section~\ref{s_hi} on this set.
Similarly, since for every $u \in S$ we maintain the set of its mates,
we can iterate over all mates of $u$.
Furthermore, for a pair of mates $u$ and $v$, we have the set of the
common 2-tight neighbors, and we can apply the randomized algorithm to this set.

Next we describe an optimization that prunes 
$(1,*)$ and $(2,*)$ moves
that are unlikely to improve the solution.
For the 
$(1,*)$ move that removes $u$, we evaluate the move only if 
the 1-tight neighborhood of $u$ changed since the last time we evaluated the
move but failed to improve the solution.
We say that the neighborhood changed
if we add $u$ to $S$ and $u$ has a non-trivial 1-tight neighborhood. 
Since our implementation of the 
$(1,*)$ move is deterministic and depends only
on the 1-tight neighborhood, we know that the move will fail.
We maintain the set $S_1$ of nodes $u \in S$ whose 1-tight neighborhood changed
but is not empty.
We pick nodes for 
$(1,*)$
moves from $S_1$.
While initializing $G_{\mathit{IN}}$, we initialize $S_1$ to include all nodes with
non-trivial 1-tight neighborhoods.
When we update $G_{\mathit{IN}}$, we also update $S_1$ 
{(see Section~\ref{s_maint}).}

For the 
$(2,*)$
move, we maintain a set $S_2$ of mate pairs $\{u,v\}$
which are eligible for the move.
We delete a pair from $S_2$ and evaluate the move that removes this
pair from $S$.
We add a pair $\{u,v\}$ to $S_2$ when they become 2-tight mates, or when
$\{u,v\}$ are 2-tight mates and their 2-tight neighborhood changes, or
when they are 2-tight mates and the 1-tight neighborhood of either $u$ or $v$
changes.
Our implementation of the 
$(2,*)$ move depends only on the 2-tight neighborhood
of the mates.
However, the implementation is randomized.
Although it is possible that one evaluation of the move succeeds and another
fails when the 2-tight neighborhood stays the same, we assume this is unlikely
and prune the move.
We maintain the set $S_2$ of mates whose 2-tight neighborhood changed.
We pick mates for 
$(2,*)$ moves from $S_2$.
While initializing $G_{\mathit{IN}}$, we 
initialize $S_2$ to all pairs of mates.
When we update $G_{\mathit{IN}}$, we update $S_2$ as well.

\subsection{AAP Moves}
For efficiency, we only look for alternating augmenting paths (AAPs) in 
the interstate graph.
The only edges on any AAP are either edges from members of $S$ to
their 1-tight and 2-tight neighbors (as edges between 2-tight mates would
not yield an alternating path).
To limit the number of AAP move evaluations,
we start a search for an AAP from a 1-tight neighbor of $v \in S_1$
($S_1$ {was introduced} in Section~\ref{s_xy}).
This way we guarantee that the move will not decrease the cardinality of
$S$, making the move more likely to succeed.
The alternating path initially contains $v$ and its single neighbor $u \in S$.
We grow the path as follows.
Let $u \in S$ be the last node on the current AAP, and let $U$
be the set of nodes on the AAP that are in $S$ 
and
$\bar{U}$ 
be the set of nodes on the AAP that are not in $S$.
We pick a mate $w$ and a 2-tight neighbor $x$ of $u$ such that
\begin{itemize}
\item
  $x$ is not a neighbor of any node of $\bar{U}$ in the input graph
  (so that the extended path will be an AAP),
\item
  neither $x$ nor $w$ are already in AAP,
\item
  the gain of flipping the extended path is maximized.
\end{itemize}
If we succeed in finding such a $\{w,x\}$ pair, we add $w$ and $x$ to the path.
Then we redefine $u$ to be $x$ and continue growing the path.
{To introduce} additional randomness, we increase the gain for every
\{$w,x\}$ pair by a random real number
$\epsilon \in [-\delta,\;\delta]$ and maximize the
perturbed gains.
We use $\delta = 50$ in our experiments.
We terminate the search if the length of the path exceeds a threshold or
the gain of flipping the path falls below a (negative) threshold.
We then perform the highest positive gain move that flips a prefix of the
final path.
If no positive gain move {is encountered,} we do nothing (the move fails).

\subsection{Maintaining the Interstate Graph}\label{s_maint}
The vast majority of the local search moves we evaluate
do not improve the solution and $G_{\mathit{IN}}$ does not change.
We need to update the graph only when a move succeeds, which happens rarely.
Our data structures speed up move evaluations and support move pruning.
The added overhead is in data structure initialization and updates.
The update complexity is non-trivial, but for sparse graphs the complexity
is much smaller than the time we save due to the improved move efficiency
and pruning.

Let $\rho(u) = |N(u) \cap S|$ denote the number of the neighbors of $u$ in $S$.
Note that for nodes $u \in S$, $\rho(u) = 0$.
We maintain $\rho(u)$ for all nodes $u \in V$.

Given an initial solution $S$, we build $G_{\mathit{IN}}$, $S_1$, and $S_2$ as follows.
We process all nodes $u \not \in S$.
For each $u$, we scan its edge list in $G$ and initialize $\rho(u)$.
If $\rho(u) = 1$, we let $N(u)\cap S = \{v\}$, add the 1-tight edge $(u,v)$
to the edge list of $u$ in $G_{\mathit{IN}}$, and add $u$ to the set 
of 1-tight neighbors of $v$.
If $\rho(u) = 2$, we let $N(u)  \cap S = \{v, w\}$, add $v$ to the set
of mates of $w$ and add $w$ to the set of mates of $v$.
We also add the pair of 2-tight edges $(u,v)$ and $(u,w)$ 
to $G_{\mathit{IN}}$.
Finally, we add $u$ to the set of 2-tight neighbors of the mates $\{v,w\}$.
We initialize $S_1$ to the set of all nodes $u \in S$ with non-empty set
of 1-tight neighbors.
We initialize $S_2$ to the set of all mate pairs $\{u,v\}$.
The initialization takes linear time.

Our algorithm updates $S$ by removing a set of nodes
$S^-$ and adding
a set of nodes $S^+$.
We break the update into a sequence of single-node updates: first we
remove nodes of $S^-$ one by one, then we add nodes of $S^+$
one by one.
We update $G_{\mathit{IN}}$ after each individual update of $S$.

After removing a node $u$ from $S$, we empty its set of 1-tight neighbors
and remove $u$ from $S_1$.
For each mate $v$ of $u$, we set the corresponding set of 2-tight
neighbors to empty and remove $u$ from the set of mates of $v$.
We also remove the pair $\{u,v\}$ from $S_2$.
Afterwards, we empty the set of mates of $u$.
We then visit its neighbors $v \in V \setminus S$.
For each neighbor $v$, we {decrement} $\rho(v)$.
We need to update $G_{\mathit{IN}}$ if $\rho(v)$ becomes zero, one, or two.

Cases for zero and two are simpler.
If the value is zero, we set the 1-tight neighbor of $v$ to null.
If the value is two, let $N(v) \cap S = \{a, b\}$.
We can find $a$ and $b$ by scanning the edge list of $v$ in $G$.
We add $a$ to the set of mates of $b$ and vice versa.
We also add $v$ to the set of 2-tight neighbors of $\{a,b\}$.
Finally, we add the 2-tight pair of edges $(v,a)$ and $(v,b)$ 
{to $G_{\mathit{IN}}$}.

If the value is one, we have to update both the old 2-tight
neighborhood and the new 1-tight neighborhood.
For the latter, we set the 1-tight neighbor of $v$ to the unique
neighbor $w \in S$, and add $v$ to the 1-tight neighbor set of $v$.
For the former update, note that $v$ was a 2-tight neighbor for mates
$\{v,w\}$ for some $w\in S$ before the removal of $v$.
We remove $v$ from the set of 2-tight neighbors of $w$ and
delete the 2-tight edge pair $(v,u)$ and $(v,w)$
{from $G_{\mathit{IN}}$}.

Now consider the addition of a node $u$ to $S$ that maintains the independence
of $S$.
We scan the edge list of $u$ and for all neighbors $v$ (guaranteed not to be
in $S$) 
{and} 
increment $\rho(v)$.
We need to update $G_{\mathit{IN}}$ if $\rho(v)$ becomes one, two, or three.

Cases for one and three are simpler.
If the value is one, we add the 1-tight edge $(v,u)$
{to $G_{\mathit{IN}}$},
add $v$ to the set
of 1-tight neighbors of $u$, and add $u$ to $S_1$.
If the value is three, $v$ has a pair of 2-tight edges $(v,a)$ and $(v,b)$,
where $a$ and $b$ are mates.
We delete $(v,a)$ and $(v,b)$
{from $G_{\mathit{IN}}$}.
Then we remove $v$ from the set of 2-tight neighbors of $a$ and $b$.
If the set becomes empty, $a$ and $b$ are no longer mates, so we remove
$a$ from the set of mates of $b$, remove $b$ from the list of mates of $a$,
and remove $\{a,b\}$ from $S_2$.

If the value is two, we have to update both the old 1-tight neighborhood
and the new 2-tight neighborhood.
For the former, let $(v,w)$ be the 1-tight edge.
We remove the edge and remove $v$ from the set of 1-tight neighbors of $w$.
If the set becomes empty, we remove $w$ from $S_1$.
In the latter case, $N(v) \cap S = \{v, w\}$ for some $w \in S$.
We add $u$ to the set of mates of $w$ and vice versa.
We also add $v$ to the set of 2-tight neighbors of $v$ and $w$.
Finally, we add $\{a,b\}$ to $S_2$.

Note that since when we add or remove $u$ to 
{or from} $S$, we may need to scan
edge lists of multiple neighbors of $u$, updating $G_{\mathit{IN}}$ when $G$ is dense
may be expensive.





\section{Experimental results}
\label{s_comp}

\subsection{Algorithms and Computational Environment}

We implemented our algorithm, which we call \emph{METAMIS}, in Java because
it is used in a production system {at Amazon}
and Java is a requirement.
For the same reason, we use doubles for node weights.
Furthermore, due to licensing restrictions, we use only standard Java
libraries.
We compiled our code using Java~8.
Our choice of programming language probably affected performance.
A C++ implementation, for example, is expected to 
be faster \citep{AlnAlHAbuHatSha12a}.

Although one can tune our algorithm for specific problem families,
we use fixed parameter settings in all experiments.

We compare our implementation to the \textit{ILSVND} algorithm
of~\cite{nogueira2018hybrid}.
The publicly available code of~\cite{nogueira2018hybrid} is implemented
in C++ and
represents weights using integers.
We made one modification {to {ILSVND}}: added the ability to warm start from an initial solution.
Given a solution in the input, we initialize the current solution of
{ILSVND} to the input solution.
We compiled ILSVND using full optimization (\texttt{-O3}) on 
an AWS r$3.4$xlarge instance \citep{r34xlarge}.

For a given instance, algorithm time-quality
plots give a lot of information about relative performance of the algorithms.
For example, one algorithm may dominate another, or one can converge to
a better solution but take longer to converge, etc.
The algorithms we compare are stochastic and algorithm performance depends
on the {pseudo-random} seed we use.
{Furthermore, the} algorithms we compare do not know if and when they reach an
optimal solution.
Usually there is a chance that a solution may improve.
However, the algorithms {\textit{converge}} in a sense that it may reach a
point of
diminishing returns when a substantial improvement becomes unlikely.
To compare the two algorithms, we put a time limit $T$ on {their
executions}.
For different problem families, the limit may be different.
We run each instance with five different pseudo-random seeds and report
the best solution value the algorithm finds.
In many cases the algorithms converge.
However, for harder problems this may take too long, and the algorithms
do not converge within the time limit.

For representative instances, we give the time-quality plots,
but we have too many instances to give all the plots.
{Therefore, we} report solution quality at {times} $T/10$ and $T/2$.
In addition, we report the time $t^*$ defined as follows.
For a given problem instance, consider the set of final solution values
over all algorithms and seed values.
Let $s$ be the smallest one of these values.
For a given algorithm, consider the run producing the best final solution
value.
For this algorithm, we define $t^*$ to be the earliest time this run
reaches the value of $s$ or higher,
Intuitively, we are comparing best runs of the algorithms being evaluated.
The choice of $s$ assures that for each algorithm, $t^*$ is well-defined.

\begin{table}
    \caption{\label{t:gr_vr} VR instances}
\vspace*{5pt}
    \begin{tabular}{l | rr | r | r}
        Graph & $|V|$ & $|E|$ & Initial Sol. & LP bound \\\hline
        MT-D-01 & \numprint{979} & \numprint{3841} & \numprint{228874404} & \numprint{238166485}\\
        MT-D-200 & \numprint{10880} & \numprint{547529} & \numprint{286750411} & \numprint{287228467}\\
        MT-D-FN & \numprint{10880} & \numprint{645026} & \numprint{290723959} & \numprint{290881566}\\
        MT-W-01 & \numprint{1006} & \numprint{3140} & \numprint{299132358} & \numprint{312121568}\\
        MT-W-200 & \numprint{12320} & \numprint{554288} & \numprint{383620215} & \numprint{384099118}\\
        MT-W-FN & \numprint{12320} & \numprint{593328} & \numprint{390596383} & \numprint{390869891}\\  
        MW-D-01 & \numprint{3988} & \numprint{19522} & \numprint{465730126} & \numprint{477563775}\\
        MW-D-20 & \numprint{10790} & \numprint{718152} & \numprint{522485254} & \numprint{531510712}\\
        MW-D-40 & \numprint{33563} & \numprint{2169909} & \numprint{533938531} & \numprint{543396252} \\
        MW-D-FN & \numprint{47504} & \numprint{4577834} & \numprint{542182073} & \numprint{549872520} \\
        MW-W-01 & \numprint{3079} & \numprint{48386} & \numprint{1268370807} & \numprint{1270311626}\\
        MW-W-05 & \numprint{10790} & \numprint{789733} & \numprint{1328552109} & \numprint{1334413294}\\
        MW-W-10 & \numprint{18023} & \numprint{2257068} & \numprint{1342415152} & \numprint{1360791627}\\
        MW-W-FN & \numprint{22316} & \numprint{3495108} & \numprint{1350675180} & \numprint{1373020454} \\ 
        MR-D-01 & \numprint{14058} & \numprint{60738} & \numprint{1664446852} & \numprint{1695332636} \\
        MR-D-03 & \numprint{21499} & \numprint{168504} & \numprint{1739544141} & \numprint{1763685757}\\
        MR-D-05 & \numprint{27621} & \numprint{295700} & \numprint{1775123794} & \numprint{1796703313}\\
        MR-D-FN & \numprint{30467} & \numprint{367408} & \numprint{1794070793} & \numprint{1809854459}\\
        MR-W-FN & \numprint{15639} & \numprint{267908} & \numprint{5386472651} & \numprint{5386842781}\\ 
        CW-T-C-1 & \numprint{266403} & \numprint{162263516} & \numprint{1298968} & \numprint{1353493}\\
        CW-T-C-2 & \numprint{194413} & \numprint{125379039} & \numprint{933792} & \numprint{957291}\\
        CW-T-D-4 & \numprint{83091} & \numprint{43680759} & \numprint{457715} & \numprint{463672}\\
        CW-T-D-6 & \numprint{83758} & \numprint{44702150} & \numprint{457605} & \numprint{463946}\\ 
        CW-S-L-1 & \numprint{411950} & \numprint{316124758} & \numprint{1622723} & \numprint{1677563}\\
        CW-S-L-2 & \numprint{443404} & \numprint{350841894} & \numprint{1692255} & \numprint{1759158}\\
        CW-S-L-4 & \numprint{430379} & \numprint{340297828} & \numprint{1709043} & \numprint{1778589}\\
        CW-S-L-6 & \numprint{267698} & \numprint{191469063} & \numprint{1159946} & \numprint{1192899}\\
        CW-S-L-7 & \numprint{127871} & \numprint{89873520} & \numprint{589723} & \numprint{599271}\\ 
        CR-T-C-1 & \numprint{602472} & \numprint{216862225} & \numprint{4605156} & \numprint{4801515} \\
        CR-T-C-2 & \numprint{652497} & \numprint{240045639} & \numprint{4844852} & \numprint{5032895}\\
        CR-T-D-4 & \numprint{651861} & \numprint{245316530} & \numprint{4789561} & \numprint{4977981}\\
        CR-T-D-6 & \numprint{381380} & \numprint{128658070} & \numprint{2953177} & \numprint{3056284}\\
        CR-T-D-7 & \numprint{163809} & \numprint{49945719} & \numprint{1451562} & \numprint{1469259}\\ 
        CR-S-L-1 & \numprint{863368} & \numprint{368431905} & \numprint{5548904} & \numprint{5768579} \\
        CR-S-L-2 & \numprint{880974} & \numprint{380666488} & \numprint{5617351} & \numprint{5867579}\\
        CR-S-L-4 & \numprint{881910} & \numprint{383405545} & \numprint{5629351} & \numprint{5869439}\\
        CR-S-L-6 & \numprint{578244} & \numprint{245739404} & \numprint{3841538} & \numprint{3990563}\\
        CR-S-L-7 & \numprint{270067} & \numprint{108503583} & \numprint{1969254} & \numprint{2041822}\\
        \hline
    \end{tabular}
\end{table}

For graph algorithms, C++ is usually faster than Java by a factor from
three to six.
We expect this to hold for our algorithm as well, especially since we make
heavy use of standard Java hash set library, which incurs significant overhead
compared to C++.
Although we do not adjust the runtimes we report, one has to keep this
in mind that if re-implemented in C++, our algorithm would be faster.

We run our experiments on an AWS r$3.4$xlarge instance with $122$GiB RAM
and $16$ virtual CPUs on Intel Xeon Ivy Bridge processors.

\subsection{Benchmark Families}

We use three problem families as benchmarks.
The first one, \emph{VR Instances} \citep{DonGolNoeParResSpa21a},
comes from our vehicle routing application.
In this application, the MWIS problem comes up in several contexts, 
and we have several instances for each of these contexts.
Table~\ref{t:gr_vr} lists the instances with their sizes.
The number of nodes in these instances ranges from 979 to 883,238;
the number of edges ranges from 3,140 to 389,304,424.
The instances are moderately sparse, but the density tends to grow with the
problem size.
The average degree is below $4$ on some small instances and over $400$ on
some large ones.

Table~\ref{t:gr_vr} also gives values for the initial solutions we use and
upper bounds on 
{optimal solution values.}
We obtain the upper bounds by solving the corresponding LP relaxation problems
to optimality.
The initial solution are good: their values are close to the upper bound.
Note that an optimal solution may not achieve the {upper} bound.

For VR instances, we have additional information: relaxed LP solutions and
initial solutions.
We use this information in practice as it yields better results.
In our experiments, we give results both for runs with and runs without initial
solutions.
We also run our algorithm with initial solutions but without
the relaxed solutions to see how much a good initial solution {matters},
and to have an apples to apples comparison with ILSVND, which
does not use this information.

\begin{table}
        \caption{\label{t:crs-gr} Problem kernels of computer,
road, and social {(CRS)}
          networks}
\vspace*{5pt}
        \begin{tabular}{l | rr}
            Graph & $|V|$ & $|E|$ \\\hline
            web-Google & \numprint{1172} & \numprint{3469} \\
            web-Stanford & \numprint{3167} & \numprint{8805} \\
            as-Skitter & \numprint{9078} & \numprint{45485} \\
            web-NotreDame & \numprint{25999} & \numprint{260681} \\
            soc-LiveJournal1 & \numprint{41917} & \numprint{240765} \\
            web-BerkStan & \numprint{71315} & \numprint{162587} \\
            roadNet-PA & \numprint{264199} & \numprint{414923} \\
            roadNet-TX & \numprint{315459} & \numprint{492722} \\
            roadNet-CA & \numprint{505103} & \numprint{787106} \\
            soc-Pokec & \numprint{906926} & \numprint{10188576} \\
                    \hline
        \end{tabular}
    \end{table}

The second problem family contains the \emph{CRS Instances}.
We derive these instances from the computer, social, and road network
instances studied by \citet{Lam19}.
The underlying graphs are natural, but the MWIS instances do not correspond to
any real application.
\citet{Lam19} describe an exact algorithm that uses local transformations
producing equivalent problems on smaller graphs.
For the Kernel instances, preprocessing based on these local transformations
reduces the graph size significantly.
The resulting graphs are the \emph{kernel} graphs.
Although the study of \citet{Lam19} focuses on exact algorithms, the authors
mention that one can combine the preprocessing with a heuristic to get
a fast heuristic algorithm.
This is what we do.
Our instances are kernel graphs for the graphs studied in \citet{Lam19}.
We use only the kernels with at {least {1000} nodes} and disregard
smaller instances.
Table~\ref{t:crs-gr} lists the CRS kernel graphs we use.
The smallest graph has 1,172 nodes and 3,469 edges, and
the {largest has} 906,926 {nodes} and 10,188,576 {edges}.
The graphs are sparse, with the average degree between $3$ and $23$.

{We remark that we tried} applying the preprocessing step to VR instances, but
the reduction in problem size was very small, below 2\%.
Therefore we do not use the preprocessing for VR instances.

One difference for our {CRS} instances is the way we assign weights to nodes.
\citeauthor{Lam19} use unweighted problem instances and assign uniform random weights.
The exact weights they use are not publicly available.
This makes it hard to reproduce the exact instances, especially when one uses a
different language platform with a different {built-in} random generator.
We assign a node the weight equal its ID modulo a constant (we use $200$),
which is reproducible.

We treat the kernel graph as an input for the algorithms we compare.
When reporting runtimes, we do not include the time to reduce a
graph to the kernel graph, and report solution values for the kernel graphs.
    \begin{table}
        \caption{\label{t:gr_osm} Map labeling instance sizes}
        \vspace*{5pt}
        \begin{tabular}{l | rrr}
            Name & $|V|$ & $|E|$ & degree \\ \hline
            florida\_AM3 & \numprint{1069} & \numprint{62088} & \numprint{62.3} \\
            alabama\_AM3 & \numprint{1614} & \numprint{117426} & \numprint{72.8}\\
            rhodeisland\_AM2 & \numprint{1103} & \numprint{81688} & \numprint{74.1}\\
            dc\_AM2 & \numprint{6360} & \numprint{592457} & \numprint{93.2}\\
            virginia\_AM3 & \numprint{3867} & \numprint{485330} & \numprint{125.5} \\
            northcarolina\_AM3 & \numprint{1178} & \numprint{189362} & \numprint{160.7}\\
            massachusetts\_AM3 & \numprint{2008} & \numprint{373537} & \numprint{186.0} \\
            kansas\_AM3 & \numprint{1605} & \numprint{408108} & \numprint{254.3}\\
            washington\_AM3 & \numprint{8030} & \numprint{2120696} & \numprint{264.1} \\
            vermont\_AM3 & \numprint{2630} & \numprint{811482} & \numprint{308.5} \\
            dc\_AM3 & \numprint{33367} & \numprint{17459296} & \numprint{523.3} \\
            oregon\_AM3 & \numprint{3670} & \numprint{1958180} & \numprint{533.6} \\
            greenland\_AM3 & \numprint{3942} & \numprint{2348539} & \numprint{595.8} \\
            idaho\_AM3 & \numprint{3208} & \numprint{2864466} & \numprint{892,9} \\
            rhodeisland\_AM3 & \numprint{13031} & \numprint{11855557} & \numprint{909.8} \\
            hawaii\_AM3 & \numprint{24436} & \numprint{40724109} & \numprint{1666.6} \\
            kentucky\_AM3 & \numprint{16871} & \numprint{54160431} & \numprint{3210.3} \\
\hline
        \end{tabular}
    \end{table}

The third problem family we use is the \emph{Map Labeling Instances}.
These are also kernel instances, but from a real map matching application,
so we treat them as a separate problem family.
We obtained the instances from Sebastian Lamm 
\citep{Lam20a}.
The motivating problem is to place a subset of labels (names of cities,
points of interest, etc.) on a map at a given magnification level
so that the labels do not intersect.
The assumption is that locations and geometry of the labels 
{are} fixed.
In the graph that models the problem, the labels correspond to nodes, and
we connect two nodes if their labels intersect.
An independent set of the labels has no overlaps.
Node weights correspond to label importance.
{See \citet{Str00}} for a discussion of unweighted version of the problem.
Higher weight subsets of the labels convey more useful information.
Table~\ref{t:gr_osm} lists the map labeling graphs we use, ordered by their
average degrees.
The number of nodes in these instances ranges from 1,068 to 33,367;
the number of edges ranges from 62,088 to 40,724,109.
Average node degrees vary from 62 to 3,210.

{Note that} VR graphs, graphs from the map labeling application
have natural cliques: For a fixed label $\ell$, nodes corresponding to
$\ell$ and the labels intersection $\ell$ form a clique.
We had access to the map labeling graph instances but not to the underlying
raw data, so we do not use clique information for this problem family.
In addition, we use kernels of the map labeling graphs.
It is unclear if the kernel-forming transformations can be adopted to
generate cliques in the kernel graph from those in the original one.

\begin{table}
    \caption{\label{t:vr-std} Results on VR instances with no additional information.}
\vspace*{5pt}
    \resizebox{\textwidth}{!}{
    \begin{tabular}{l | rrr | r | rrr | r}
        & \multicolumn{4}{c}{METAMIS} & \multicolumn{4}{c}{ILSVND}\\
        Name & $w_{10\%}$ & $w_{50\%}$ & $w$ & $t^*[s]$& $w_{10\%}$ & $w_{50\%}$ & $w$ & $t^*[s]$ \\\hline
        MT-D-01&\textbf{\numprint{238166485}}&\textbf{\numprint{238166485}}&\textbf{\numprint{238166485}}&\textbf{\numprint{0.948}}&\textbf{\numprint{238166485}}&\textbf{\numprint{238166485}}&\textbf{\numprint{238166485}}&\numprint{1.290}\\
        MT-D-200 & \textbf{\numprint{286976422}}&\textbf{\numprint{287048909}}&\textbf{\numprint{287048909}}&\textbf{\numprint{188.1}}&\numprint{286838210}&\numprint{286838210}&\numprint{286943799}&\numprint{2276}\\
        MT-D-FN &\textbf{\numprint{290866943}}&\textbf{\numprint{290866943}}&\textbf{\numprint{290866943}}&\textbf{\numprint{104.4}}&\numprint{290393532}&\numprint{290666380}&\numprint{290666380}&\numprint{561.6}\\
        MT-W-01 & \textbf{\numprint{312121568}}&\textbf{\numprint{312121568}}&\textbf{\numprint{312121568}}&\numprint{0.278}&\textbf{\numprint{312121568}}&\textbf{\numprint{312121568}}&\textbf{\numprint{312121568}}&\textbf{\numprint{0.080}}\\
        MT-W-200 & \numprint{383818136}&\textbf{\numprint{383961099}}&\textbf{\numprint{383961323}}&\numprint{1433}&\textbf{\numprint{383865836}}&\numprint{383896403}&\numprint{383896403}&\textbf{\numprint{1036}}\\
        MT-W-FN & \numprint{390688944}&\textbf{\numprint{390830057}}&\textbf{\numprint{390854593}}&\textbf{\numprint{568.1}}&\textbf{\numprint{390715890}}&\numprint{390798842}&\numprint{390798842}&\numprint{709.2}\\
        MW-D-01 &\textbf{\numprint{476099262}}&\textbf{\numprint{476164209}}&\textbf{\numprint{476334711}}&\textbf{\numprint{267.9}}&\numprint{475653439}&\numprint{475906790}&\numprint{475906790}&\numprint{1173}\\
        MW-D-20 & \textbf{\numprint{524255389}}&\textbf{\numprint{525036493}}&\textbf{\numprint{525124575}}&\textbf{\numprint{85.40}}&\numprint{520854115}&\numprint{522415092}&\numprint{523138978}&\numprint{2685}\\
        MW-D-40 & \textbf{\numprint{533934442}}&\textbf{\numprint{535707479}}&\textbf{\numprint{536520199}}&\textbf{\numprint{81.36}}&\numprint{530227261}&\numprint{532272896}&\numprint{532400878}&\numprint{1830}\\
        MW-D-FN & \textbf{\numprint{539754400}}&\textbf{\numprint{541372345}}&\textbf{\numprint{541918916}}&\textbf{\numprint{98.34}}&\numprint{532663872}&\numprint{537238784}&\numprint{537674129}&\numprint{2466}\\
        MW-W-01 & \textbf{\numprint{1270305952}}&\textbf{\numprint{1270305952}}&\textbf{\numprint{1270305952}}&\textbf{\numprint{0.500}}&\numprint{1246949460}&\numprint{1246949460}&\numprint{1246949460}&\numprint{23.66}\\
        MW-W-05 & \textbf{\numprint{1328958047}}&\textbf{\numprint{1328958047}}&\textbf{\numprint{1328958047}}&\textbf{\numprint{19.96}}&\numprint{1327687399}&\numprint{1328707787}&\numprint{1328707787}&\numprint{984.8}\\
        MW-W-10 &\textbf{\numprint{1340878388}}&\textbf{\numprint{1342899725}}&\textbf{\numprint{1342899725}}&\textbf{\numprint{1204}}&\numprint{1331002512}&\numprint{1341482310}&\numprint{1342067985}&\numprint{1876}\\
        MW-W-FN & \textbf{\numprint{1349369736}}&\textbf{\numprint{1350818543}}&\textbf{\numprint{1350818543}}&\textbf{\numprint{527.7}}&\numprint{1334835589}&\numprint{1348128240}&\numprint{1350159705}&\numprint{3584}\\
        MR-D-01 & \textbf{\numprint{1689074331}}&\textbf{\numprint{1689520690}}&\textbf{\numprint{1689781114}}&\textbf{\numprint{15.52}}&\numprint{1683529331}&\numprint{1686091786}&\numprint{1687842856}&\numprint{2906}\\
        MR-D-03 & \textbf{\numprint{1753188475}}&\textbf{\numprint{1753968167}}&\textbf{\numprint{1754110286}}&\textbf{\numprint{20.34}}&\numprint{1743429914}&\numprint{1747269072}&\numprint{1749972580}&\numprint{3257}\\
        MR-D-05 & \textbf{\numprint{1784519403}}&\textbf{\numprint{1785664042}}&\textbf{\numprint{1786342921}}&\textbf{\numprint{19.56}}&\numprint{1770832093}&\numprint{1774407092}&\numprint{1777876780}&\numprint{3595}\\
        MR-D-FN & \textbf{\numprint{1795912642}}&\textbf{\numprint{1797284091}}&\textbf{\numprint{1797573192}}&\textbf{\numprint{22.65}}&\numprint{1779897201}&\numprint{1785545729}&\numprint{1788331878}&\numprint{3388}\\
        MR-W-FN & \textbf{\numprint{5357026363}}&\numprint{5358386615}&\numprint{5358386615}&\numprint{1442}&\numprint{5352347338}&\textbf{\numprint{5370471580}}&\textbf{\numprint{5371649721}}&\textbf{\numprint{461.6}}\\
        CW-T-C-1 & \textbf{\numprint{1310223}}&\textbf{\numprint{1315122}}&\textbf{\numprint{1317775}}&\textbf{\numprint{94.52}}&\numprint{1290974}&\numprint{1299279}&\numprint{1302478}&\numprint{3585}\\
        CW-T-C-2 &\textbf{\numprint{924664}}&\textbf{\numprint{929626}}&\textbf{\numprint{931802}}&\textbf{\numprint{189.7}}&\numprint{914736}&\numprint{921021}&\numprint{922858}&\numprint{3599}\\
        CW-T-C-4 & \textbf{\numprint{454769}}&\textbf{\numprint{456565}}&\textbf{\numprint{457185}}&\textbf{\numprint{324.4}}&\numprint{452035}&\numprint{453741}&\numprint{454544}&\numprint{2365}\\
        CW-T-D-6 & \textbf{\numprint{455823}}&\textbf{\numprint{457382}}&\textbf{\numprint{457790}}&\textbf{\numprint{70.48}}&\numprint{452366}&\numprint{454254}&\numprint{454254}&\numprint{1582}\\
        CW-S-L-1 & \textbf{\numprint{1623280}}&\textbf{\numprint{1630417}}&\textbf{\numprint{1634950}}&\textbf{\numprint{261.9}}&\numprint{1603051}&\numprint{1615247}&\numprint{1620756}&\numprint{3597}\\
        CW-S-L-2 &\textbf{\numprint{1695131}}&\textbf{\numprint{1704424}}&\textbf{\numprint{1708820}}&\textbf{\numprint{225.3}}&\numprint{1670836}&\numprint{1685870}&\numprint{1690536}&\numprint{3596}\\
        CW-S-L-4 & \textbf{\numprint{1712553}}&\textbf{\numprint{1722542}}&\textbf{\numprint{1725591}}&\textbf{\numprint{173.7}}&\numprint{1689318}&\numprint{1701309}&\numprint{1706264}&\numprint{3599}\\
        CW-S-L-6 & \textbf{\numprint{1150229}}&\textbf{\numprint{1156916}}&\textbf{\numprint{1158925}}&\textbf{\numprint{138.4}}&\numprint{1136356}&\numprint{1142720}&\numprint{1145694}&\numprint{3086}\\
        CW-S-L-7 & \textbf{\numprint{582925}}&\textbf{\numprint{585929}}&\textbf{\numprint{587288}}&\textbf{\numprint{125.2}}&\numprint{577087}&\numprint{581583}&\numprint{581583}&\numprint{1278}\\
        CR-T-C-1 & \textbf{\numprint{4617204}}&\textbf{\numprint{4644635}}&\textbf{\numprint{4654419}}&\textbf{\numprint{58.16}}&\numprint{4508901}&\numprint{4558780}&\numprint{4576695}&\numprint{3598}\\
        CR-T-C-2 & \textbf{\numprint{4834040}}&\textbf{\numprint{4863054}}&\textbf{\numprint{4874346}}&\textbf{\numprint{62.29}}&\numprint{4715023}&\numprint{4772847}&\numprint{4789909}&\numprint{3600}\\
        CR-T-D-4 & \textbf{\numprint{4778868}}&\textbf{\numprint{4808490}}&\textbf{\numprint{4817281}}&\textbf{\numprint{56.91}}&\numprint{4663588}&\numprint{4716258}&\numprint{4734674}&\numprint{3598}\\
        CR-T-D-6 &\textbf{\numprint{2945721}}&\textbf{\numprint{2964007}}&\textbf{\numprint{2970011}}&\textbf{\numprint{94.09}}&\numprint{2896260}&\numprint{2921540}&\numprint{2929671}&\numprint{3574}\\
        CR-T-D-7 & \textbf{\numprint{1431915}}&\textbf{\numprint{1438896}}&\textbf{\numprint{1440281}}&\textbf{\numprint{148.4}}&\numprint{1411061}&\numprint{1423279}&\numprint{1426400}&\numprint{3581}\\
        CR-S-L-1 & \textbf{\numprint{5547038}}&\textbf{\numprint{5575602}}&\textbf{\numprint{5588489}}&\textbf{\numprint{72.42}}&\numprint{5400658}&\numprint{5464532}&\numprint{5487254}&\numprint{3595}\\
        CR-S-L-2 & \textbf{\numprint{5652928}}&\textbf{\numprint{5680688}}&\textbf{\numprint{5691892}}&\textbf{\numprint{57.91}}&\numprint{5491814}&\numprint{5561766}&\numprint{5586973}&\numprint{3580}\\
        CR-S-L-4 & \textbf{\numprint{5634886}}&\textbf{\numprint{5671369}}&\textbf{\numprint{5681336}}&\textbf{\numprint{65.09}}&\numprint{5477340}&\numprint{5550943}&\numprint{5572856}&\numprint{3573}\\
        CR-S-L-6 & \textbf{\numprint{3833391}}&\textbf{\numprint{3851432}}&\textbf{\numprint{3859513}}&\textbf{\numprint{92.45}}&\numprint{3751019}&\numprint{3793995}&\numprint{3808314}&\numprint{3599}\\
        CR-S-L-7 & \textbf{\numprint{1977161}}&\textbf{\numprint{1986354}}&\textbf{\numprint{1989879}}&\textbf{\numprint{90.90}}&\numprint{1940573}&\numprint{1957872}&\numprint{1963579}&\numprint{3584}\\
\hline
    \end{tabular}
    }
\end{table}

\subsection{{Results for} VR Family}

For VR instances, we run experiments on the standard version of the problem
and on the augmented version, where we have a good initial solution and
clique information, as we do in our application.
We set the time limit T = \numprint{3600} seconds.

Table~\ref{t:vr-std} gives results for the standard version.
For each instance in the table, column $w_{10\%}$ shows the best solution 
value found at time point $T/10$, column $w_{50\%}$  
shows the best solution value found at time point $T/2$, and column $w$
shows the best solution value found when the process is finished at time $T$.
METAMIS finds better solutions than ILSVND except for three instances.
For two instances, MT-D-01 and MT-W-01, solution quality is the same.
On MW-W-01, the ILSVND solution is better, but only by $0.8\%$.
All three exceptions happen on smaller instances and both algorithms converge
quickly. There is no improvement after time $T/10$.

An interesting observation is that on MT-D-01, MT-W-01 and MT-W-FN,
solution values match the corresponding upper bounds given in
Table~\ref{t:gr_vr},
so the solutions are optimal.
Since the upper bound need not be tight, it is possible that we solve other
instances to optimality, but do not have a proof.

On larger instances, METAMIS has better final values as well as
better values at times $T/10$ and $T/2$.
On the problem with the highest number of nodes, CR-S-L-3, the difference in
the final values is $2.1\%$.
Note that on large instances, neither algorithms converged in time $T$.

\begin{table}[t]
\centering
    \caption{\label{t:vr-aug-1} METAMIS+LP results on VR instances.}
\vspace*{3pt}
{\normalsize
    \begin{tabular}{l | rrr | r }
        Name & $w_{10\%}$ & $w_{50\%}$ & $w$ & $t^*[s]$ \\\hline\hline
MT-D-01 & \textbf{\numprint{238166485}}&\textbf{\numprint{238166485}}&\textbf{\numprint{238166485}}&\textbf{\numprint{0.109}}\\
MT-D-200 & \textbf{\numprint{287038328}}&\textbf{\numprint{287048081}}&\textbf{\numprint{287048081}}&\textbf{\numprint{69.51}}\\
MT-D-FN & \textbf{\numprint{290771450}}&\textbf{\numprint{290771450}}&\textbf{\numprint{290771450}}&$-$\\
MT-W-01 & \textbf{\numprint{312121568}}&\textbf{\numprint{312121568}}&\textbf{\numprint{312121568}}&\numprint{0.122}\\
MT-W-200 & \textbf{\numprint{383971124}}&\numprint{383985408}&\numprint{383985408}&\textbf{\numprint{893.0}}\\
MT-W-FN & \textbf{\numprint{390828160}}&\textbf{\numprint{390869891}}&\textbf{\numprint{390869891}}&\textbf{\numprint{139.9}}\\
MW-D-01 & \textbf{\numprint{475886356}}&\textbf{\numprint{475987082}}&\textbf{\numprint{475987082}}&\textbf{\numprint{270.2}} \\
MW-D-20 & \textbf{\numprint{525052532}}&\textbf{\numprint{525402318}}&\textbf{\numprint{525486034}}&\numprint{8.694} \\
MW-D-40 & \textbf{\numprint{535705687}}&\textbf{\numprint{536210247}}&\textbf{\numprint{536735155}}&\textbf{\numprint{0.434}}\\
MW-D-FN & \textbf{\numprint{543098071}}&\textbf{\numprint{543622238}}&\textbf{\numprint{543857187}}&$-$\\
MW-W-01 & \numprint{1269314742}&\textbf{\numprint{1269344846}}&\textbf{\numprint{1269344846}}&\numprint{672.0}\\
MW-W-05 & \textbf{\numprint{1328958047}}&\textbf{\numprint{1328958047}}&\textbf{\numprint{1328958047}}&\textbf{\numprint{0.431}}\\
MW-W-10 & \textbf{\numprint{1342915691}}&\textbf{\numprint{1342915691}}&\textbf{\numprint{1342915691}}&\textbf{\numprint{0.511}}\\
MW-W-FN & \textbf{\numprint{1350814699}}&\numprint{1350814699}&\textbf{\numprint{1350818543}}&$-$\\
MR-D-01 & \textbf{\numprint{1688024106}}&\textbf{\numprint{1688777944}}&\textbf{\numprint{1689278470}}&\textbf{\numprint{7.245}}\\
MR-D-03 & \textbf{\numprint{1756186736}}&\textbf{\numprint{1756989875}}&\textbf{\numprint{1757227519}}&\textbf{\numprint{5.123}}\\
MR-D-05 &\textbf{\numprint{1787220357}}&\textbf{\numprint{1787666207}}&\textbf{\numprint{1787849777}}&\textbf{\numprint{19.91}}\\
MR-D-FN & \textbf{\numprint{1798215807}}&\textbf{\numprint{1798926794}}&\textbf{\numprint{1799452160}}&\textbf{\numprint{17.40}}\\
MR-W-FN &\textbf{\numprint{5386842781}}&\textbf{\numprint{5386842781}}&\textbf{\numprint{5386842781}}&\textbf{\numprint{0.503}}\\
&&&&\\
CW-T-C-1 & \textbf{\numprint{1334884}}&\textbf{\numprint{1336953}}&\textbf{\numprint{1338064}}&\numprint{30.69}\\
CW-T-C-2 & \textbf{\numprint{944404}}&\textbf{\numprint{945748}}&\textbf{\numprint{945886}}&\textbf{\numprint{25.86}}\\
CW-T-D-4 &\textbf{\numprint{460643}}&\textbf{\numprint{461027}}&\textbf{\numprint{461056}}&\numprint{2.000}\\
CW-T-D-6 &\textbf{\numprint{460982}}&\textbf{\numprint{461223}}&\textbf{\numprint{461312}}&\numprint{2.717}\\
CW-S-L-1 & \textbf{\numprint{1656404}}&\textbf{\numprint{1660475}}&\textbf{\numprint{1660815}}&\textbf{\numprint{46.27}}\\
CW-S-L-2 & \textbf{\numprint{1731077}}&\textbf{\numprint{1735964}}&\textbf{\numprint{1738128}}&\textbf{\numprint{85.11}}\\
CW-S-L-4 & \textbf{\numprint{1748029}}&\textbf{\numprint{1752354}}&\textbf{\numprint{1753803}}&\numprint{91.08}\\
CW-S-L-6 & \numprint{1174005}&\textbf{\numprint{1175931}}&\textbf{\numprint{1177156}}&\textbf{\numprint{27.48}}\\
CW-S-L-7 & \numprint{593045}&\textbf{\numprint{593744}}&\numprint{593891}&\textbf{\numprint{4.825}}\\
CR-T-C-1 & \textbf{\numprint{4730533}}&\textbf{\numprint{4739684}}&\textbf{\numprint{4743040}}&\textbf{\numprint{17.92}}\\
CR-T-C-2 &\textbf{\numprint{4954613}}&\textbf{\numprint{4966121}}&\textbf{\numprint{4968952}}&\numprint{25.80}\\
CR-T-D-4 &\textbf{\numprint{4898377}}&\textbf{\numprint{4908285}}&\textbf{\numprint{4911646}}&\numprint{19.69}\\
CR-T-D-6 &\textbf{\numprint{3017902}}&\textbf{\numprint{3022448}}&\textbf{\numprint{3024523}}&\textbf{\numprint{28.67}}\\
CR-T-D-7 &\textbf{\numprint{1458949}}&\textbf{\numprint{1459958}}&\textbf{\numprint{1460240}}&\numprint{16.88}\\
CR-S-L-1 &\textbf{\numprint{5672398}}&\textbf{\numprint{5686939}}&\textbf{\numprint{5692891}}&\numprint{36.79}\\
CR-S-L-2 & \textbf{\numprint{5763866}}&\textbf{\numprint{5780859}}&\textbf{\numprint{5784034}}&\numprint{24.90}\\
CR-S-L-4 &\textbf{\numprint{5756016}}&\textbf{\numprint{5771410}}&\textbf{\numprint{5777081}}&\textbf{\numprint{24.08}}\\
CR-S-L-6 & \textbf{\numprint{3926517}}&\textbf{\numprint{3933476}}&\textbf{\numprint{3936137}}&\numprint{22.24}\\
CR-S-L-7 &\textbf{\numprint{2014584}}&\textbf{\numprint{2018371}}&\textbf{\numprint{2019428}}&\textbf{\numprint{34.09}}\\
    \hline
    \end{tabular}
    }
\end{table}

\begin{table}[t]
\centering
    \caption{\label{t:vr-aug-2} 
        METAMIS
results on VR instances.}
\vspace*{3pt}
{\normalsize
    \begin{tabular}{l | rrr | r }
        Name & $w_{10\%}$ & $w_{50\%}$ & $w$ & $t^*[s]$ \\\hline\hline
MT-D-01 
&\textbf{\numprint{238166485}}&\textbf{\numprint{238166485}}&\textbf{\numprint{238166485}}&\numprint{0.373} \\
MT-D-200 &
\numprint{287010847}&\numprint{287018324}&\numprint{287036715}&\numprint{122.6}\\
MT-D-FN & 
\numprint{290752054}&
\textbf{\numprint{290771450}}&
\textbf{\numprint{290771450}}&$-$\\
MT-W-01 &
\textbf{\numprint{312121568}}&\textbf{\numprint{312121568}}&\textbf{\numprint{312121568}}&\numprint{0.320}\\
MT-W-200 &
\numprint{383804298}&\textbf{\numprint{383986483}}&\textbf{\numprint{383986483}}&\numprint{1343} \\
MT-W-FN 
&\numprint{390787880}&\numprint{390848998}&\numprint{390856179}&\numprint{710.1}
\\
MW-D-01 
&\numprint{475549969}&\numprint{475814986}&\numprint{475955989}&\numprint{2278} \\
MW-D-20 
&\numprint{524574519}&\numprint{525068939}&\numprint{525192291}&\textbf{\numprint{7.699}} \\
MW-D-40 
&\numprint{535436892}&\numprint{535711417}&\numprint{536092070}&\numprint{0.474} \\
MW-D-FN 
&\numprint{542740347}&\numprint{543253226}&\numprint{543374394}&$-$\\
MW-W-01 
&\textbf{\numprint{1269344846}}&\textbf{\numprint{1269344846}}&\textbf{\numprint{1269344846}}&\textbf{\numprint{0.603}} \\
MW-W-05 
&\textbf{\numprint{1328958047}}&\textbf{\numprint{1328958047}}&\textbf{\numprint{1328958047}}&\numprint{0.447} \\
MW-W-10 
&\textbf{\numprint{1342915691}}&\textbf{\numprint{1342915691}}&\textbf{\numprint{1342915691}}&\numprint{1.255} \\
MW-W-FN 
&\numprint{1350771010}&\textbf{\numprint{1350818542}}&\textbf{\numprint{1350818543}}&$-$ \\
MR-D-01 
&\numprint{1687486503}&\numprint{1687807619}&\numprint{1688118984}&\numprint{16.84} \\
MR-D-03 
&\numprint{1755768835}&\numprint{1756154528}&\numprint{1756337669}&\numprint{12.31} \\
MR-D-05 
&\numprint{1786084687}&\numprint{1786734327}&\numprint{1786755817}&\numprint{73.22} \\
MR-D-FN 
&\numprint{1798075911}&\numprint{1798571155}&\numprint{1798661823}&\numprint{38.60} \\
MR-W-FN 
&\textbf{\numprint{5386842781}}&\textbf{\numprint{5386842781}}&\textbf{\numprint{5386842781}}&\numprint{0.855} \\
&&&&\\
CW-T-C-1 
&\numprint{1333129}&\numprint{1335297}&\numprint{1336563}&\textbf{\numprint{22.44}} \\
CW-T-C-2 
&\numprint{943366}&\numprint{944785}&\numprint{945565}&\numprint{27.72} \\
CW-T-D-4 
&\numprint{460554}&\numprint{460852}&\numprint{461025}&\textbf{\numprint{1.828}} \\
CW-T-D-6 
&\numprint{460815}&\numprint{461057}&\numprint{461174}&\textbf{\numprint{2.706}} \\
CW-S-L-1 
&\textbf{\numprint{1656404}}&\textbf{\numprint{1660475}}&\textbf{\numprint{1660815}}&\numprint{90.11} \\
CW-S-L-2 
&\numprint{1730208}&\numprint{1734736}&\numprint{1736245}&\numprint{109.3} \\
CW-S-L-4 
&\numprint{1746941}&\numprint{1751474}&\numprint{1751988}&\textbf{\numprint{84.05}} \\
CW-S-L-6 
&\textbf{\numprint{1174169}}&\numprint{1175886}&\numprint{1176233}&\numprint{33.79} \\
CW-S-L-7 
&\textbf{\numprint{593077}}&\textbf{\numprint{593744}}&\textbf{\numprint{593947}}&\numprint{6.622} \\
CR-T-C-1 
&\numprint{4725855}&\numprint{4735644}&\numprint{4738289}&\numprint{18.10} \\
CR-T-C-2 
&\numprint{4950818}&\numprint{4962045}&\numprint{4964446}&\textbf{\numprint{19.83}} \\
CR-T-D-4 
&\numprint{4896504}&\numprint{4906792}&\numprint{4909999}&\textbf{\numprint{17.86}} \\
CR-T-D-6 
&\numprint{3016890}&\numprint{3022046}&\numprint{3023349}&\numprint{40.35} \\
CR-T-D-7 
&\numprint{1458571}&\numprint{1459653}&\numprint{1459948}&\textbf{\numprint{8.115}} \\
CR-S-L-1 
&\numprint{5668764}&\numprint{5685884}&\numprint{5690515}&\textbf{\numprint{21.79}} \\
CR-S-L-2 
&\numprint{5759512}&\numprint{5775002}&\numprint{5780449}&\textbf{\numprint{22.53}} \\
CR-S-L-4 
&\numprint{5755282}&\numprint{5771391}&\numprint{5775704}&\numprint{24.18} \\
CR-S-L-6 
&\numprint{3923574}&\numprint{3932059}&\numprint{3935089}&\textbf{\numprint{19.00}} \\
CR-S-L-7 
&\numprint{2013466}&\numprint{2017034}&\numprint{2017836}&\numprint{40.24} \\
    \hline
    \end{tabular}
    }
\end{table}

\begin{table}[t]
\centering
    \caption{\label{t:vr-aug-3} ILSVND results on VR instances.}
\vspace*{3pt}
{\normalsize
    \begin{tabular}{l | rrr | r }
        Name & $w_{10\%}$ & $w_{50\%}$ & $w$ & $t^*[s]$ \\\hline\hline
MT-D-01 
&\textbf{\numprint{238166485}}&\textbf{\numprint{238166485}}&\textbf{\numprint{238166485}}&\numprint{1.473}\\
MT-D-200 
&\numprint{286949274}&\numprint{286973561}&\numprint{286973561}&\numprint{363.1}\\
MT-D-FN 
&\numprint{290723959}&\numprint{290723959}&\numprint{290723959}&$-$\\
MT-W-01 
&\textbf{\numprint{312121568}}&\textbf{\numprint{312121568}}&\textbf{\numprint{312121568}}&\textbf{\numprint{0.063}}\\
MT-W-200 
&\numprint{383808376}&\numprint{383979962}&\numprint{383979962}&\numprint{1721}\\
MT-W-FN 
&\numprint{390805960}&\numprint{390805960}&\numprint{390805960}&\numprint{196.2}\\
MW-D-01 
&\numprint{475523699}&\numprint{475732519}&\numprint{475825497}&\numprint{2134}\\
MW-D-20 
&\numprint{523248884}&\numprint{523248884}&\numprint{523248884}&\numprint{26.98}\\
MW-D-40 
&\numprint{534040009}&\numprint{534040009}&\numprint{534040009}&\numprint{7.797}\\
MW-D-FN 
&\numprint{542182073}&\numprint{542182073}&\numprint{542182073}&$-$\\
MW-W-01 
&\textbf{\numprint{1269344846}}&\textbf{\numprint{1269344846}}&\textbf{\numprint{1269344846}}&\numprint{1.247}\\
MW-W-05 
&\numprint{1328955871}&\numprint{1328955871}&\numprint{1328955871}&\numprint{4.266}\\
MW-W-10 
&\numprint{1342847887}&\numprint{1342847887}&\numprint{1342847887}&\numprint{19.39}\\
MW-W-FN 
&\numprint{1350675180}&\numprint{1350675180}&\numprint{1350675180}&$-$\\
MR-D-01 
&\numprint{1684211854}&\numprint{1686046636}&\numprint{1686452467}&\numprint{2763}\\
MR-D-03 
&\numprint{1751006933} &\numprint{1752345436}&\numprint{1752769459}&\numprint{3305}\\
MR-D-05 
&\numprint{1782046226}&\numprint{1782560957}&\numprint{1783836981}&\numprint{3525}\\
MR-D-FN 
&\numprint{1794949819}&\numprint{1794949819}&\numprint{1796037791}&\numprint{3564}\\
MR-W-FN 
&\numprint{5386838669}&\numprint{5386838669}&\numprint{5386838669}&\numprint{10.01}\\
&&&&\\
CW-T-C-1 
&\numprint{1322410}&\numprint{1326551}&\numprint{1327556}&\numprint{3501}\\
CW-T-C-2 
&\numprint{939568}&\numprint{940356}&\numprint{940356}&\numprint{701.3}\\
CW-T-D-4 
&\numprint{458360}&\numprint{458360}&\numprint{458360}&\numprint{48.65}\\
CW-T-D-6 
&\numprint{459096}&\numprint{459096}&\numprint{459096}&\numprint{80.73}\\
CW-S-L-1 
&\numprint{1644241}&\numprint{1649006}&\numprint{1651483}&\numprint{3585}\\
CW-S-L-2 
&\numprint{1714923}&\numprint{1722672}&\numprint{1724930}&\numprint{3452}\\
CW-S-L-4 
&\numprint{1733007}&\numprint{1739992}&\numprint{1742459}&\numprint{3553}\\
CW-S-L-6 
&\numprint{1167611}&\numprint{1169914}&\numprint{1170096}&\numprint{1886}\\
CW-S-L-7 
&\numprint{591398}&\numprint{591398}&\numprint{591398}&\numprint{161.2}\\
CR-T-C-1 
&\numprint{4665849}&\numprint{4687422}&\numprint{4696568}&\numprint{3591}\\
CR-T-C-2 
&\numprint{4891697}&\numprint{4912140}&\numprint{4920058}&\numprint{3585}\\
CR-T-D-4 
&\numprint{4836312}&\numprint{4859311}&\numprint{4867272}&\numprint{3597}\\
CR-T-D-6 
&\numprint{2990174}&\numprint{2999852}&\numprint{3004067}&\numprint{3593}\\
CR-T-D-7 
&\numprint{1455226}&\numprint{1456048}&\numprint{1456048}&\numprint{752.0}\\
CR-S-L-1 
&\numprint{5590089}&\numprint{5617916}&\numprint{5630437}&\numprint{3596}\\
CR-S-L-2 
&\numprint{5670522}&\numprint{5701371}&\numprint{5715430}&\numprint{3589}\\
CR-S-L-4 
&\numprint{5676163}&\numprint{5701271}&\numprint{5715256}&\numprint{3598}\\
CR-S-L-6 
&\numprint{3883092}&\numprint{3898898}&\numprint{3905831}&\numprint{3597}\\
CR-S-L-7 
&\numprint{1998320}&\numprint{2006129}&\numprint{2007794}&\numprint{3488}\\
    \hline
    \end{tabular}
    }
\end{table}

Tables~\ref{t:vr-aug-1}, \ref{t:vr-aug-2}, and \ref{t:vr-aug-3}
shows results for the VR instances for METAMIS+LP, METAMIS, and ILSVND, 
respectively.
On the instances where ILSVND does not find any improvement, $t^*$ is undefined.
While both algorithms allow a warm start from a given solution,
the METAMIS+LP version of our algorithm uses clique information to compute
the relaxed LP solution, and uses it to guide local search.,
We evaluate both versions of METAMIS to see how much the LP relaxation helps.
As in the case of no initial solution, 
the algorithms converge on most of the small instances and
do not converge on larger instances.

\begin{figure}[tbh]
    \centering
    \includegraphics[width=.9\textwidth]{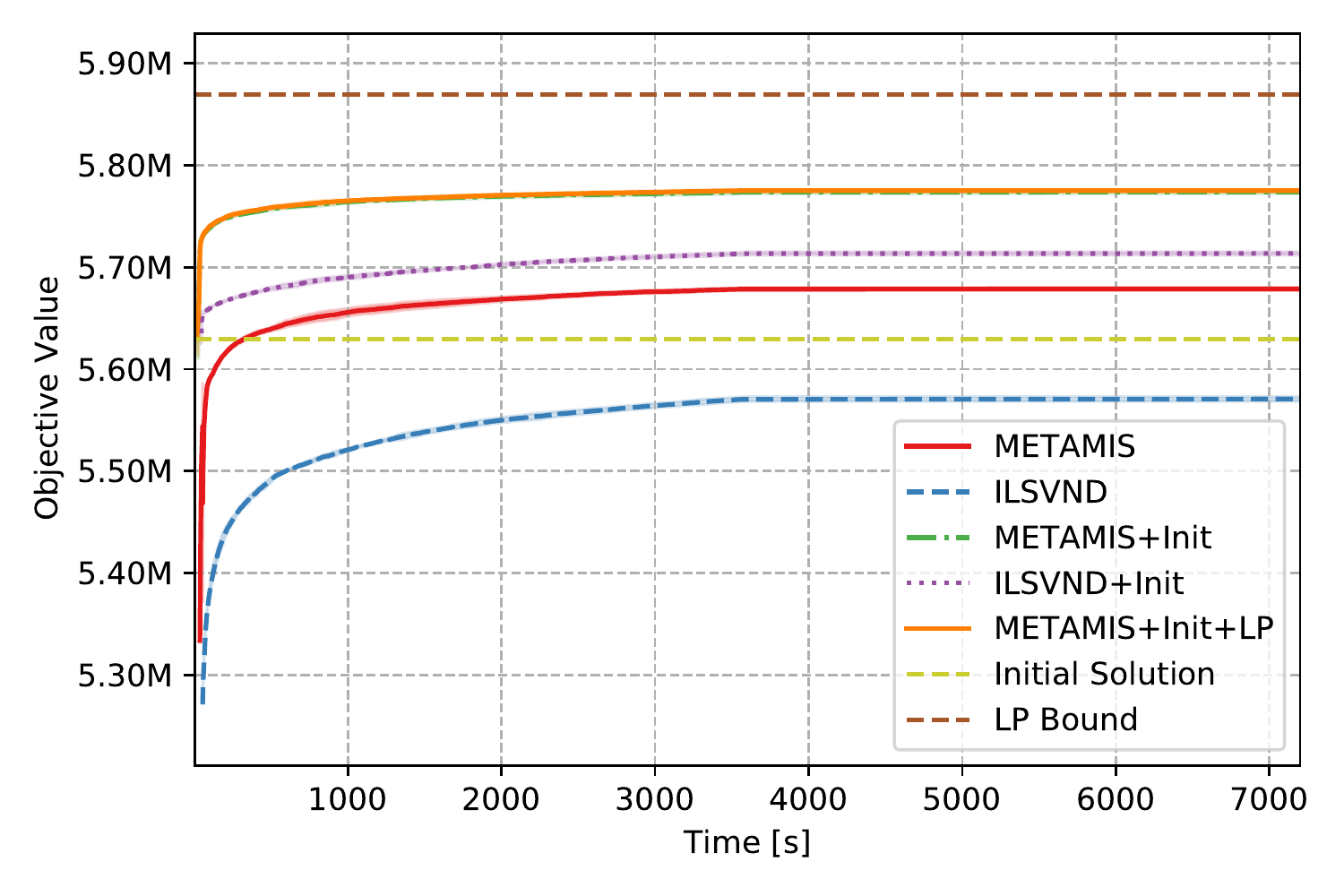}
    \caption{Time-quality plot for the standard CR-S-L-4 instance
      with $95\%$ confidence intervals, initial solution and LP bound.
Note that the plots for METAMIS+Init and METAMIS+Init+LP are very close}
    \label{f:vr}
\end{figure}

Recall that with no initial solution, we found optimal solutions for
MT-D-01 and 
{MT-W-01.}
With the initial solution, METAMIS+LP finds an optimal solution for two
more instances, MT-W-FN and MR-W-FN.
METAMIS finds an optimal solution for the latter instance, but not for the
former.
ILSVND does not find any new optimal solutions.

Next we discuss the effect of a good initial solution, comparing results
for METAMIS and ILSVND from Tables~\ref{t:vr-std} and~\ref{t:vr-aug-1}-\ref{t:vr-aug-3}.
Comparing initial solution values from Table~\ref{t:gr_vr} with solutions
obtained by solving the problems from scratch, we see that
in many cases, the initial solution is better than
the solution computed from scratch.
In fact, for ILSVND, most solutions are worse than the corresponding
initial solution.
This confirms that our initial solutions are good.

With the warm start, both variants of our algorithm, METAMIS and META\-MIS+\-LP,
dominate ILSVND, producing same or (in most cases) better quality solutions.
ILSVND is also slower on all instances except one.

To evaluate the benefit of using LP relaxation, we compare METAMIS+LP
with METAMIS.
On most instances, METAMIS+LP dominates METAMIS.
The latter never finds a better solution.
For about $1/3$ of the instances, solution quality is the same, and
for the remaining $2/3$, METAMIS+LP performs better.
The same holds for intermediate times $T/10$ and $T/2$ except for one
instance at $T/2$ where METAMIS solution value is slightly better.

Plots help visualize relative algorithm performance.
We plot the data for $2$-hour runs of all algorithms on one of the largest 
instances in {Figure}~\ref{f:vr}.
Neither algorithm converges. Both with and without an initial solution, METAMIS dominates ILSVND.
With the initial solution, the algorithms converge to a better value
and dominate the corresponding algorithms without an initial solution.
Comparing METAMIS+LP with METAMIS, we see that the relaxed solution
helps. The former algorithm dominates the latter.

 \begin{table}
   \caption{\label{t:crs} CRS family results}
\vspace*{5pt}
     \resizebox{\textwidth}{!}{
     \begin{tabular}{l | rrr | r | rrr | r}
         & \multicolumn{4}{c}{METAMIS} & \multicolumn{4}{c}{ILSVND}\\
         Name & $w_{10\%}$ & $w_{50\%}$ & $w$ & $t^*[s]$& $w_{10\%}$ & $w_{50\%}$ & $w$ & $t^*[s]$ \\\hline
     web-Google & \textbf{\numprint{50143}}&\textbf{\numprint{50143}}&\textbf{\numprint{50143}}&\numprint{0.398}&\textbf{\numprint{50143}}&\textbf{\numprint{50143}}&\textbf{\numprint{50143}}&\textbf{\numprint{0.105}}\\
         web-Stanford & \textbf{\numprint{140528}}&\textbf{\numprint{140528}}&\textbf{\numprint{140528}}&\textbf{\numprint{1.263}}&\numprint{140524}&\textbf{\numprint{140528}}&\textbf{\numprint{140528}}&\numprint{627.5}\\
         as-Skitter & \textbf{\numprint{403178}}&\textbf{\numprint{403220}}&\numprint{403243}&\numprint{1443}&\numprint{403100}&\numprint{403159}&\textbf{\numprint{403254}}&\textbf{\numprint{1254}}\\
         web-NotreDame & \textbf{\numprint{882498}}&\textbf{\numprint{883034}}&\textbf{\numprint{883073}}&\textbf{\numprint{98.10}}&\numprint{882347}&\numprint{882347}&\numprint{882347}&\numprint{131.0}\\
         soc-LiveJournal1 & \textbf{\numprint{1639058}}&\textbf{\numprint{1639150}}&\textbf{\numprint{1639150}}&\textbf{\numprint{58.56}}&\numprint{1638224}&\numprint{1638537}&\numprint{1638619}&\numprint{1467}\\
         web-BerkStan & \textbf{\numprint{3202724}}&\textbf{\numprint{3203853}}&\textbf{\numprint{3203896}}&\textbf{\numprint{50.64}}&\numprint{3192742}&\numprint{3196624}&\numprint{3196865}&\numprint{1478}\\
         roadNet-PA & \textbf{\numprint{12327884}}&\textbf{\numprint{12381128}}&\textbf{\numprint{12386969}}&\textbf{\numprint{909.7}}&\numprint{12165765}&\numprint{12330319}&\numprint{12342736}&\numprint{7143}\\
         roadNet-TX & \textbf{\numprint{14706445}}&\textbf{\numprint{14798448}}&\textbf{\numprint{14809215}}&\textbf{\numprint{1323}}&\numprint{14466017}&\numprint{14726128}&\numprint{14758276}&\numprint{7193}\\
         roadNet-CA & \textbf{\numprint{23447525}}&\textbf{\numprint{23688219}}&\textbf{\numprint{23731691}}&\textbf{\numprint{1601}}&\numprint{22837850}&\numprint{23400764}&\numprint{23576536}&\numprint{7200}\\
         soc-Pokec & \textbf{\numprint{40093941}}&\textbf{\numprint{40154755}}&\textbf{\numprint{40167652}}&\textbf{\numprint{119.5}}&\numprint{39193450}&\numprint{39677478}&\numprint{39882658}&\numprint{7200}\\
     \hline
     \end{tabular}
     }
 \end{table}

\subsection{{Results for} CRS Problems}

 Table~\ref{t:crs} gives results for the CRS family.
 On this family we set the time limit to 1500 seconds for all problems
 except for the four largest ones, {as the algorithms {did not}
 converge on them in 1500 seconds.}
For the four largest problems, we set the limit to 7200 seconds.

\begin{figure}[tbh]
    \centering
    \includegraphics[width=.9\textwidth]{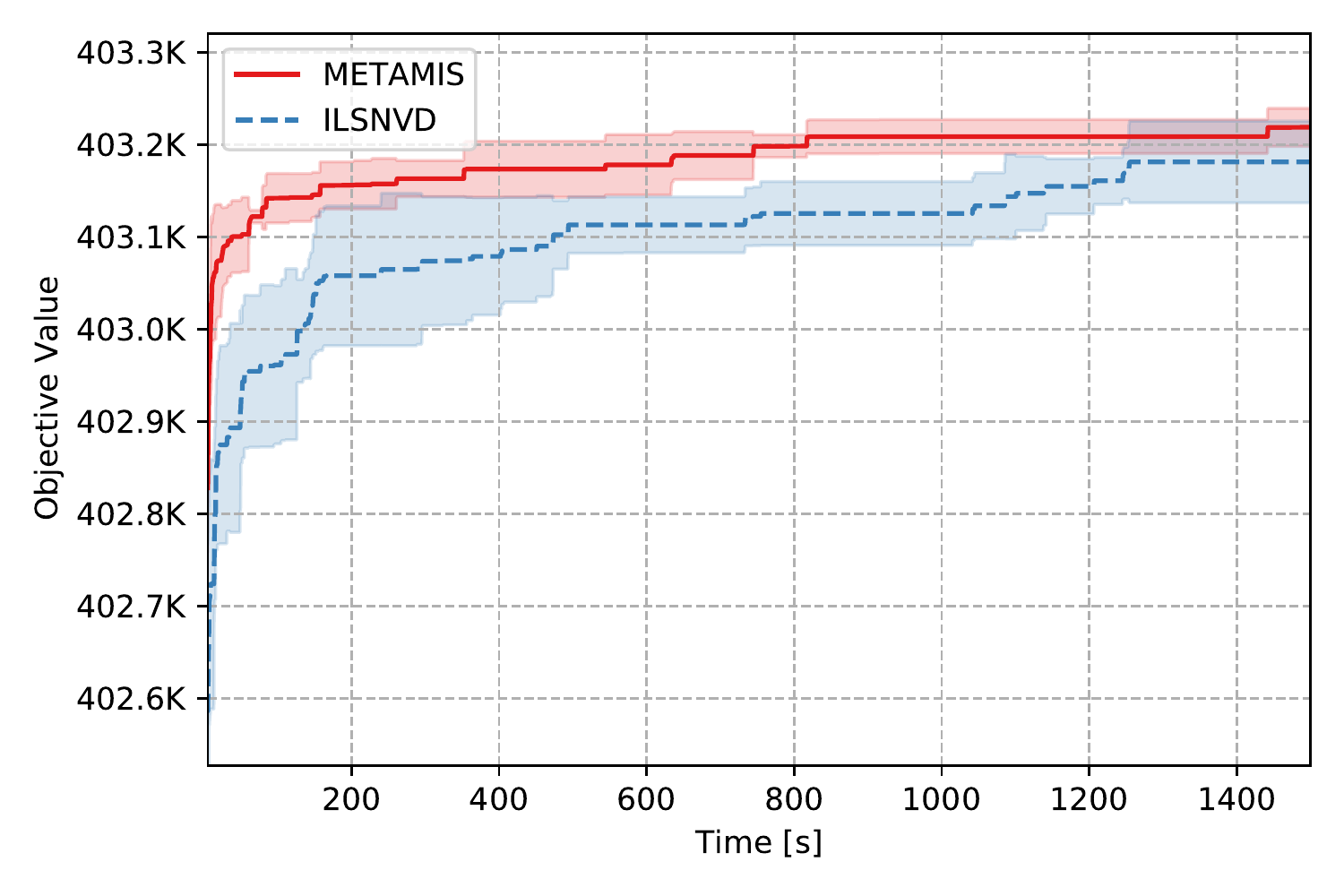}
    \caption{Time-quality plot for as-Skitter with $95\%$ confidence intervals}
    \label{f:as-skitter}
\end{figure}

\begin{figure}[tbh]
    \centering
    \includegraphics[width=.9\textwidth]{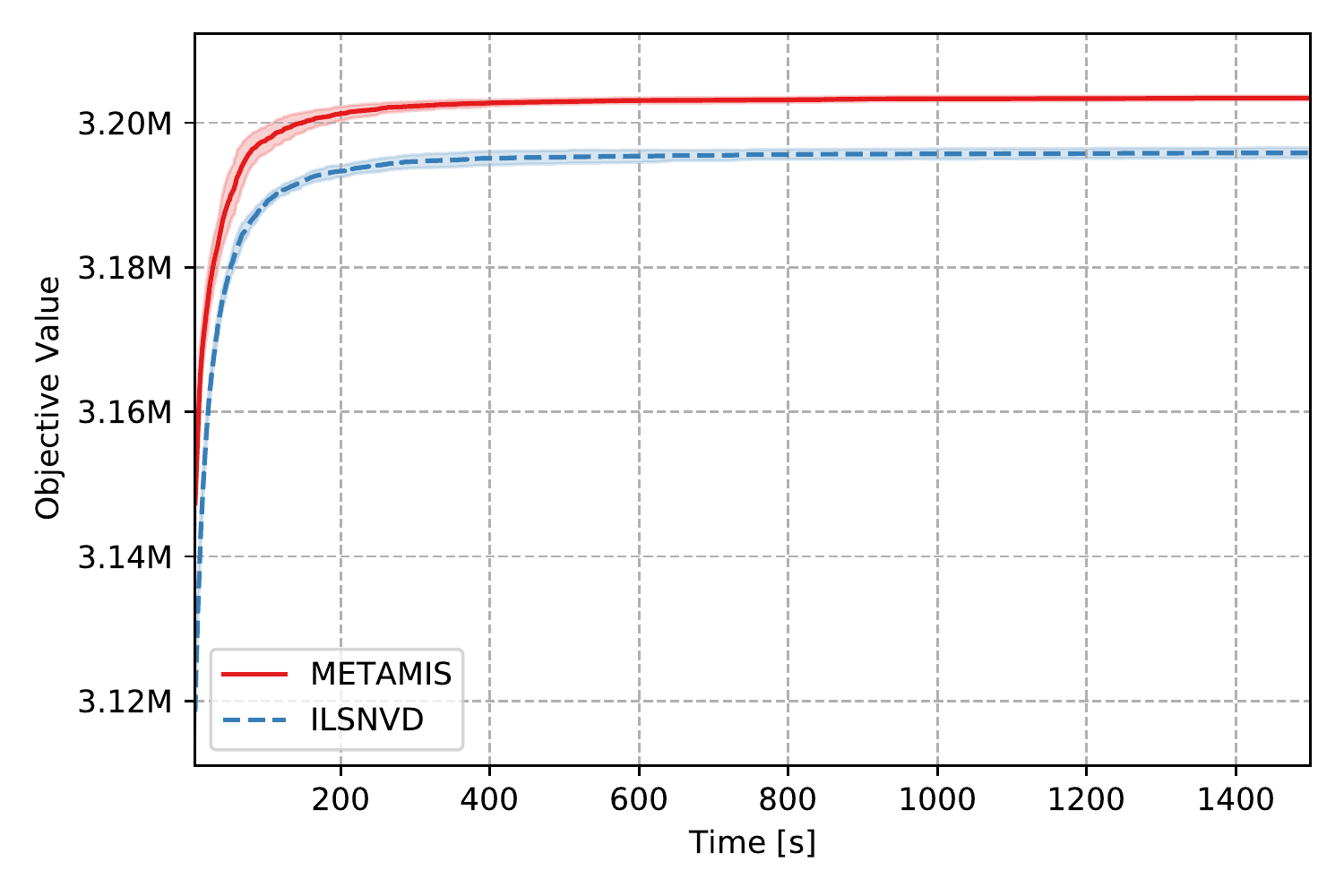}
    \caption{Time-quality plot for web-BerkSt with $95\%$ confidence intervals}
    \label{f:berk-st}
\end{figure}

\begin{figure}[tbh]
    \centering
    \includegraphics[width=.9\textwidth]{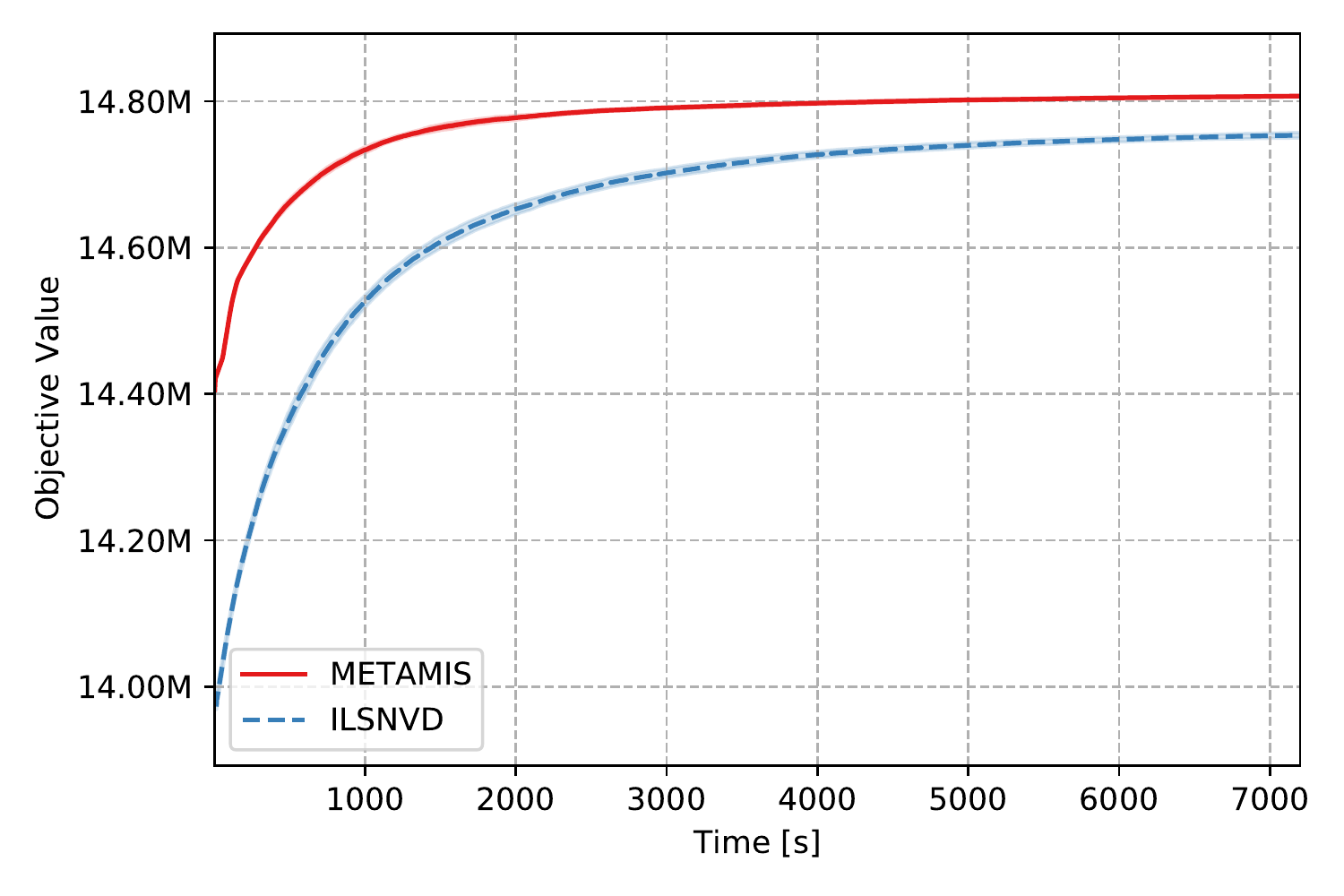}
    \caption{Time-quality plot for roadNet-TX with $95\%$ confidence intervals}
    \label{f:road-tx}
\end{figure}

On web-Google, both algorithms find solutions with the same value in
under a second and fail to improve it afterwards (the solution may be optimal).
On web-Stanford and as-Skitter, ILSVND performs worse 
{on average} and has higher variance.
The best solution of ILSVND is the same for web-Stanford and slightly better for
as-Skitter {(see Figure~\ref{f:as-skitter})}.
On three other smaller instances, METAMIS finds a better solution and
dominates ILSVND.
For example, see the plot for web-BerkSt {in} Figure~\ref{f:berk-st}.

On the four larger instances METAMIS dominates ILSVND and converges to better values.
See, for example, Figure~\ref{f:road-tx} for a plot of {two}
hour-long runs on RoadNet-TX.

\subsection{{Results for} 
{Map Labeling Problems}
}

 \begin{table}
   \caption{\label{t:osm} Map Labeling family results.}
\vspace*{5pt}
   \resizebox{\textwidth}{!}{
   \begin{tabular}{l | rrr | r | rrr | r}
     & \multicolumn{4}{c}{METAMIS} & \multicolumn{4}{c}{ILSVND}\\
     Name & $w_{10\%}$ & $w_{50\%}$ & $w$ & $t^*[s]$& $w_{10\%}$ & $w_{50\%}$ & $w$ & $t^*[s]$ \\\hline
     florida\_AM3 &\textbf{\numprint{46132}}&\textbf{\numprint{46132}}&\textbf{\numprint{46132}}&\numprint{3.080}&\textbf{\numprint{46132}}&\textbf{\numprint{46132}}&\textbf{\numprint{46132}}&\textbf{\numprint{0.024}}\\
     alabama\_AM3 & \textbf{\numprint{45449}}&\textbf{\numprint{45449}}&\textbf{\numprint{45449}}&\numprint{5.412}&\textbf{\numprint{45449}}&\textbf{\numprint{45449}}&\textbf{\numprint{45449}}&\textbf{\numprint{0.104}}\\
     rhodeisland\_AM2 &\textbf{\numprint{43722}}&\textbf{\numprint{43722}}&\textbf{\numprint{43722}}&\numprint{0.269}&\textbf{\numprint{43722}}&\textbf{\numprint{43722}}&\textbf{\numprint{43722}}&\textbf{\numprint{0.048}}\\
     dc\_AM2 &\textbf{\numprint{100302}}&\textbf{\numprint{100302}}&\textbf{\numprint{100302}}&\textbf{\numprint{58.38}}&\textbf{\numprint{100302}}&\textbf{\numprint{100302}}&\textbf{\numprint{100302}}&\numprint{107.3}\\
     virginia\_AM3 &\textbf{\numprint{97873}}&\textbf{\numprint{97873}}&\textbf{\numprint{97873}}&\numprint{9.080}&\textbf{\numprint{97873}}&\textbf{\numprint{97873}}&\textbf{\numprint{97873}}&\textbf{\numprint{2.750}}\\
     northcarolina\_AM3 &\textbf{\numprint{13062}}&\textbf{\numprint{13062}}&\textbf{\numprint{13062}}&\numprint{0.376}&\textbf{\numprint{13062}}&\textbf{\numprint{13062}}&\textbf{\numprint{13062}}&\textbf{\numprint{0.040}}\\
     massachusetts\_AM3 &\textbf{\numprint{17224}}&\textbf{\numprint{17224}}&\textbf{\numprint{17224}}&\numprint{2.223}&\textbf{\numprint{17224}}&\textbf{\numprint{17224}}&\textbf{\numprint{17224}}&\textbf{\numprint{0.142}}\\
     kansas\_AM3 &\textbf{\numprint{5694}}&\textbf{\numprint{5694}}&\textbf{\numprint{5694}}&\numprint{3.383}&\textbf{\numprint{5694}}&\textbf{\numprint{5694}}&\textbf{\numprint{5694}}&\textbf{\numprint{0.209}}\\
     washington\_AM3 &\textbf{\numprint{118196}}&\textbf{\numprint{118196}}&\textbf{\numprint{118196}}&\textbf{\numprint{62.35}}&\textbf{\numprint{118196}}&\textbf{\numprint{118196}}&\textbf{\numprint{118196}}&\numprint{119.8}\\
     vermont\_AM3 &\textbf{\numprint{28349}}&\textbf{\numprint{28349}}&\textbf{\numprint{28349}}&\numprint{9.332}&\textbf{\numprint{28349}}&\textbf{\numprint{28349}}&\textbf{\numprint{28349}}&\textbf{\numprint{0.297}}\\
     dc\_AM3 &\numprint{141785}&\numprint{142736}&\numprint{142910}&\numprint{1347}&\textbf{\numprint{142917}}&\textbf{\numprint{143014}}&\textbf{\numprint{143014}}&\textbf{\numprint{111.8}}\\
     oregon\_AM3 &\textbf{\numprint{34471}}&\textbf{\numprint{34471}}&\textbf{\numprint{34471}}&\textbf{\numprint{11.56}}&\textbf{\numprint{34471}}&\textbf{\numprint{34471}}&\textbf{\numprint{34471}}&\numprint{49.42}\\
     greenland\_AM3 &\textbf{\numprint{11960}}&\textbf{\numprint{11960}}&\textbf{\numprint{11960}}&\textbf{\numprint{28.02}}&\numprint{11959}&\textbf{\numprint{11960}}&\textbf{\numprint{11960}}&\numprint{178.4}\\
     idaho\_AM3 &\textbf{\numprint{9224}}&\textbf{\numprint{9224}}&\textbf{\numprint{9224}}&\textbf{\numprint{21.23}}&\textbf{\numprint{9224}}&\textbf{\numprint{9224}}&\textbf{\numprint{9224}}&\numprint{127.4}\\
     rhodeisland\_AM3 &\numprint{80897}&\textbf{\numprint{81013}}&\textbf{\numprint{81013}}&\numprint{449.7}&\textbf{\numprint{80980}}&\numprint{80980}&\numprint{80980}&\textbf{\numprint{101.7}}\\
     hawaii\_AM3 &\numprint{58394}&\textbf{\numprint{58808}}&\textbf{\numprint{58819}}&\numprint{1207}&\textbf{\numprint{58779}}&\numprint{58783}&\numprint{58814}&\textbf{\numprint{757.0}}\\
     kentucky\_AM3 &\numprint{30736}&\numprint{30777}&\numprint{30789}&\numprint{1387}&\textbf{\numprint{31084}}&\textbf{\numprint{31101}}&\textbf{\numprint{31103}}&\textbf{\numprint{11.90}}\\
   \hline
   \end{tabular}
   }
 \end{table}

Table~\ref{t:osm} gives our results for the Map Labeling instances.
On these instances, we ran the two algorithms for $1,500$ seconds.
The quality of the solutions the algorithms find is the same except
for four instances.
METAMIS is better on rhodeisland\_AM3 and hawaii\_AM3, and
ILSVND -- on dc\_AM3 and kentucky\_AM3. 
Although both algorithms converge to a similar value, ILSVND
usually converges faster.

\begin{figure}[tbh]
    \centering
    \includegraphics[width=.9\textwidth]{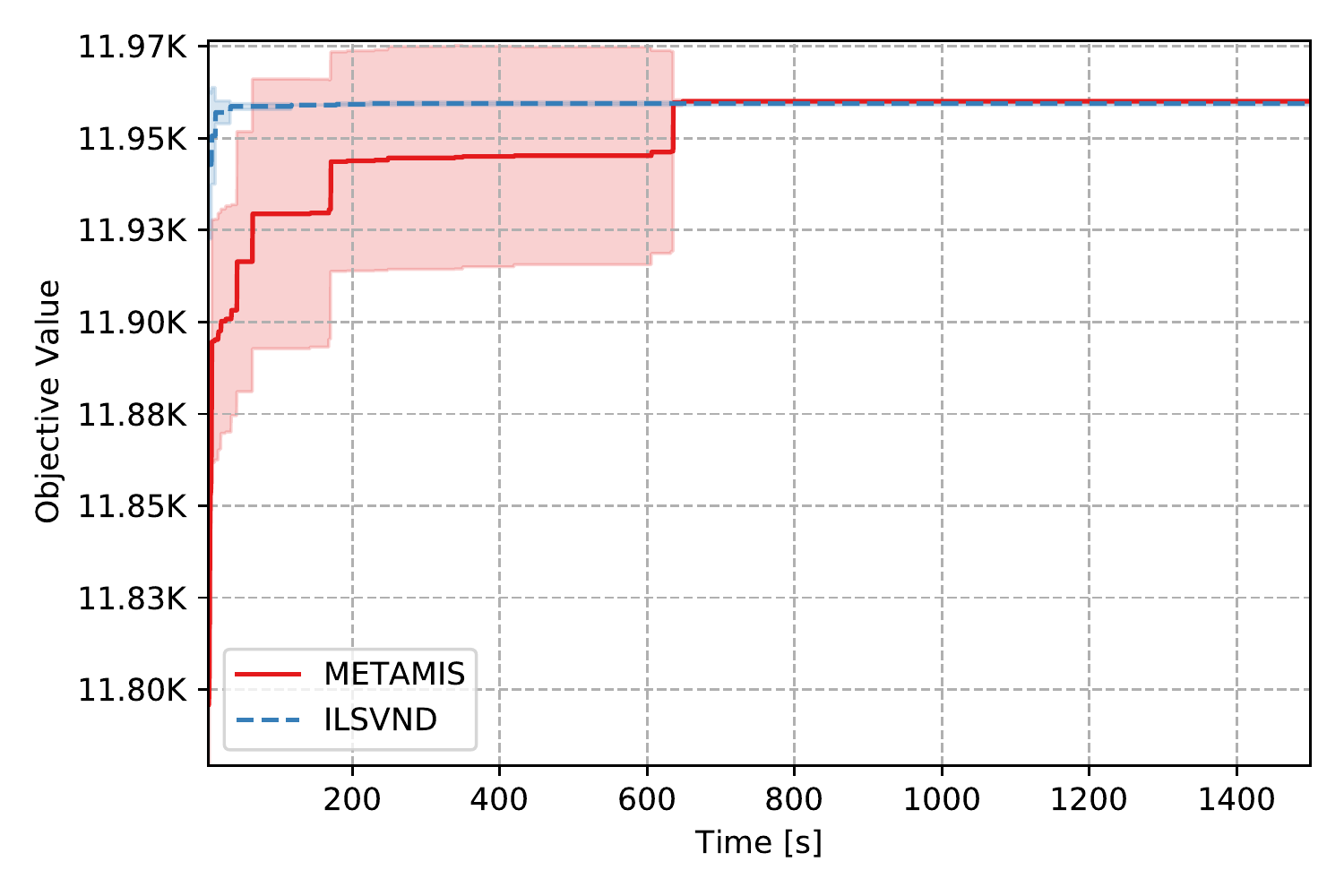}
    \caption{Time-quality plot for greenland\_AM3
      with $95\%$ confidence intervals}
    \label{f:greenland}
\end{figure}

\begin{figure}[h]
    \centering
    \includegraphics[width=.9\textwidth]{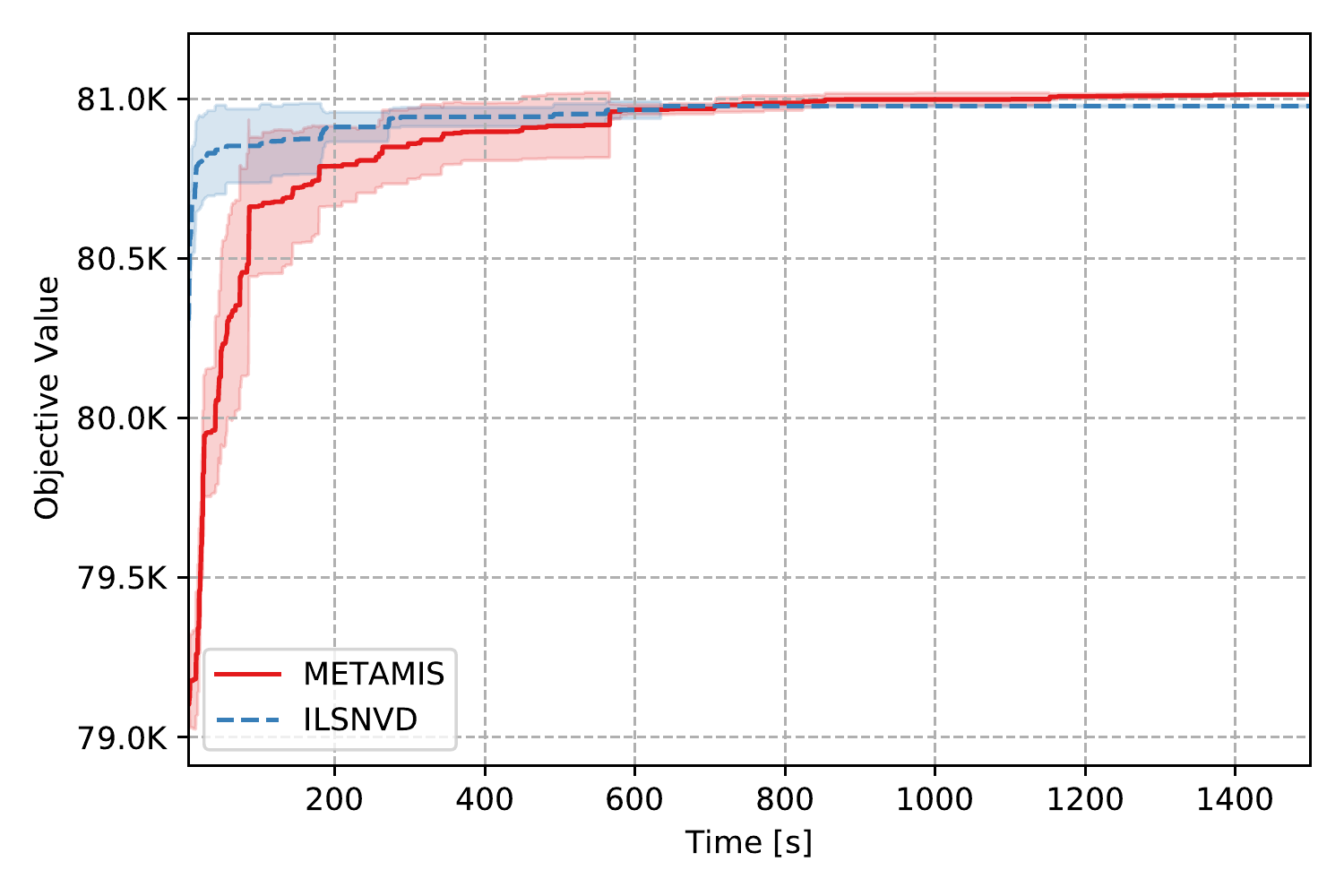}
    \caption{Time-quality plot for rhodeisland\_AM3
      with $95\%$ confidence intervals}
    \label{f:rhodeisland}
\end{figure}

\begin{figure}[tbh]
    \centering
    \includegraphics[width=.9\textwidth]{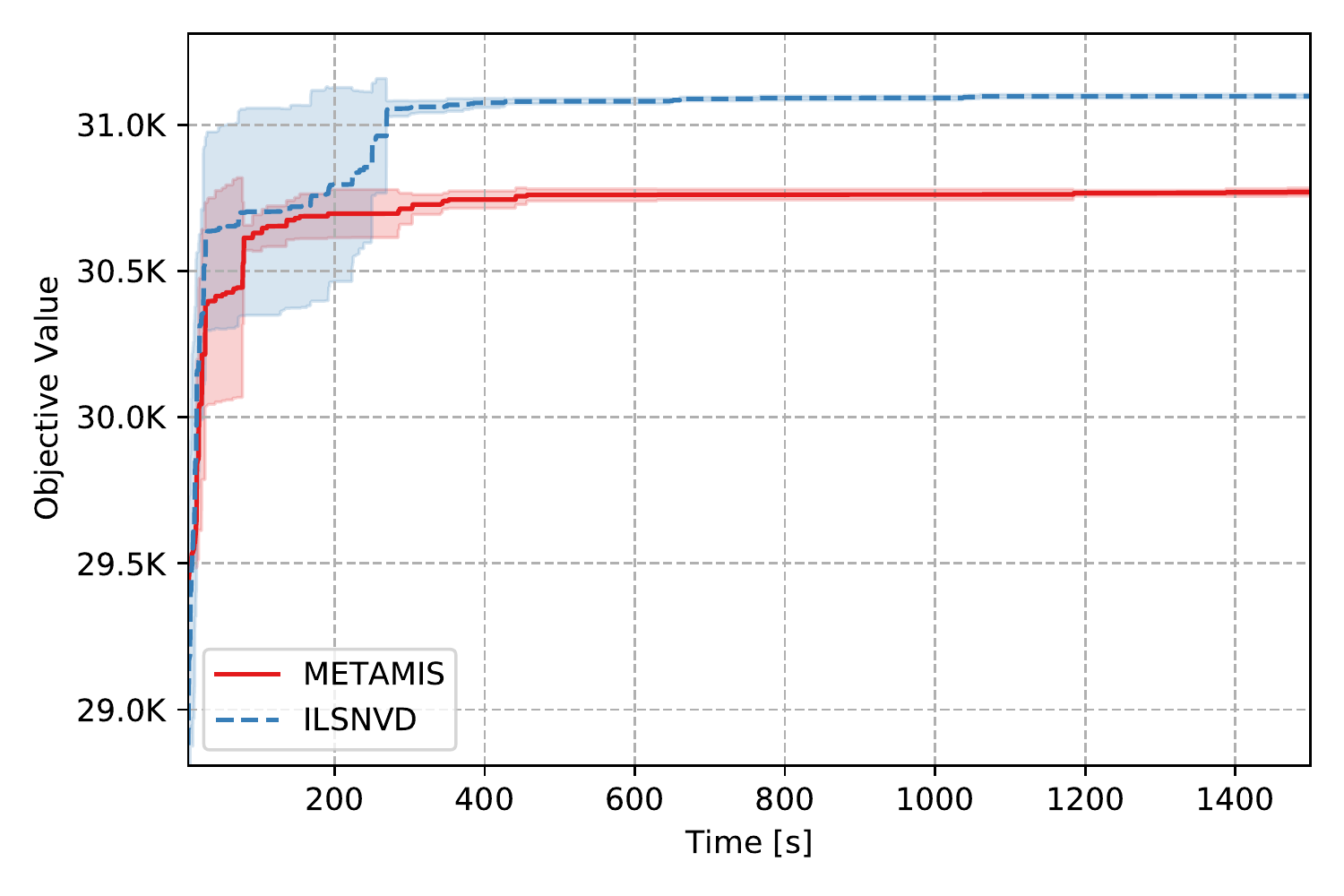}
    \caption{Time-quality plot for kentucky\_AM3 with $95\%$ confidence intervals}
    \label{f:kentucky}
\end{figure}

Figure~\ref{f:greenland} illustrates convergence to an essentially the same
value using greenland\_AM3 as an example.
Figure~\ref{f:rhodeisland} gives the plot for rhodeisland\_AM3, where
the picture is similar but METAMIS converges to {a} slightly better value.
Figure~\ref{f:kentucky} gives the plot for kentucky\_AM3, where
ILSVND converges to a better value.


\section{Concluding remarks}
\label{s_concl}

We developed METAMIS for a real-{world} VR application for which
even a small improvement in solution quality yields substantial cost reduction.
  We published benchmark VR instances in~\citet{DonGolNoeParResSpa21a,DonGolNoeParResSpa21b}.
  These instances are structurally different from other MWIS benchmarks and include large instances.
  Our study is the first to include the new benchmark, and we show that METAMIS works well on the VR instances.
METAMIS is also competitive on {CRS} and Map Matching instances.

METAMIS uses a more sophisticated set of local search moves and introduces
data structures that facilitate efficient implementation of these moves.
We also introduce a new variation of path relinking tailored to large problems.
In addition, we show how to use a good relaxed solution to guide local search.
These techniques add to the metaheuristic design toolset.
  We hope that our ideas will lead to even more efficient MWIS algorithms. 
The ideas may also prove useful in methaheuristic algorithms for other problems. 

We do not include DIMACS problem instances~\citep{JT96} in our experiments.
These instances have a special structure that is an artifact of how
the instances are generated.
The instances originate from clique instances, both synthetic and real-life.
One takes a compliment of a graph to obtain the corresponding unweighted
maximum independent set instances.
To get MWIS instances, one adds weights, which are either
random or ID modulo a constant.
The original clique graphs are sparse, and the resulting MWIS graphs are 
very dense.
Even for the clique instances that correspond to real-life applications,
the resulting MWIS instances do not.
These problems are relatively small and easy.
Many have been solved exactly~\citep{nogueira2018hybrid}.

Preliminary experiments show that our algorithm performs worse than ILSVND
on DIMACS instances.
We tuned our code for large VR instances and used the same parameter values
in all experiments.
A different choice of parameter values may improve performance on
specific problem families, such as the DIMACS family {\citep{Kummer2020,KumResSou2020a}}.
Also, for dense graphs, a different (e.g., {adjacency} matrix) representation
of the input graph may be more efficient.

For the Map Labeling problem family, the fact that on most instances
METAMIS and ILSVND converge to the same solution value suggests that
in many cases the solutions may be optimal.
One way to {verify} this conjecture is to prove optimality by using
local search solutions to warm start an exact solver.
Even if the solver {cannot solve} such a problem from scratch, it may be able
to prove optimality of a given solution.

\bibliographystyle{plainnat}
\bibliography{bibliography,graspbib}   

\end{document}